%% file: ChestXRay.tex
\documentclass[runningheads,a4paper]{llncs}
\usepackage{hyperref}
\usepackage{amssymb}
\usepackage{graphicx,epstopdf,epsfig,subcaption,amsmath}
\usepackage{tikz}
\usepackage{setspace}
\usepackage{color, colortbl}
\usepackage{hhline}
\usepackage{url}
\usepackage{multirow}
\usepackage{array,booktabs}

\urldef{\mailsa}\path|{alfred.hofmann, ursula.barth, ingrid.haas, frank.holzwarth,|
\urldef{\mailsb}\path|anna.kramer, leonie.kunz, christine.reiss, nicole.sator,|
\urldef{\mailsc}\path|erika.siebert-cole, peter.strasser, lncs}@springer.com|

\usepackage{floatrow}
\newfloatcommand{capbtabbox}{table}[][\FBwidth]

\newsavebox\CBox
\def\textBF#1{\sbox\CBox{#1}\resizebox{\wd\CBox}{\ht\CBox}{\textbf{#1}}}
\definecolor{Gray}{gray}{0.9}

\begin{document}
\mainmatter  

\makeatletter
\def\@normalsize{\@setsize\normalsize{10pt}\xpt\@xpt
\abovedisplayskip 10pt plus2pt minus5pt\belowdisplayskip
\abovedisplayskip \abovedisplayshortskip \z@
plus3pt\belowdisplayshortskip 6pt plus3pt
minus3pt\let\@listi\@listI}
\def\subsize{\@setsize\subsize{12pt}\xipt\@xipt}
\def\section{\@startsection {section}{1}{\z@}{1.0ex plus
1ex minus .2ex}{.2ex plus .2ex}{\large\bf}}
\def\subsection{\@startsection {subsection}{2}{\z@}{.2ex
plus 1ex} {.2ex plus .2ex}{\subsize\bf}} \makeatother

\newcommand{\Section}[1]{\section{\hskip -1em.~~#1}}
\newcommand{\SubSection}[1]{\subsection{\hskip -1em.~~#1}}
\def\@listI{%
 \leftmargin\leftmargini
 \partopsep 0pt
 \parsep 0pt
 \topsep 0pt
 \itemsep pt
 \relax
} \long\def\@makecaption#1#2{
 \vskip -5pt
 \setbox\@tempboxa\hbox{\small{#1\,:\,#2}}
  \ifdim \wd\@tempboxa >\hsize \unhbox\@tempboxa\par \else
  \hbox to\hsize{\hfil\box\@tempboxa\hfil}
\fi \vskip -0.2cm}

\jot=0pt \abovedisplayskip=3pt \belowdisplayskip=3pt
\abovedisplayshortskip=0pt \belowdisplayshortskip=0pt

\newcommand{\etal}{\textit{et al}.}
\newcommand{\ie}{\textit{i}.\textit{e}.}
\newcommand{\eg}{\textit{e}.\textit{g}.}

\title{Iterative Attention Mining for Weakly Supervised Thoracic Disease Pattern Localization in Chest X-Rays}

\author{Jinzheng Cai\inst{1} \and Le Lu\inst{2} \and Adam P. Harrison\inst{2} \and Xiaoshuang Shi\inst{1} \and \\ Pingjun Chen\inst{1} \and Lin Yang\inst{1} \\
	\institute{
		University of Florida, Gainesville, FL, 32611, USA
		\and AI-Infra, NVIDIA Corp, Bethesda, MD, 20814, USA
		\\
		\email{jimmycai@ufl.edu}
	}
}

\titlerunning{Iterative Attention Mining}
\authorrunning{J. Cai~\etal{}}

%
%
%


%
%

\toctitle{Lecture Notes in Computer Science}
\tocauthor{MICCAI submission \# 246}
\maketitle

\begin{abstract}
	Given image labels as the only supervisory signal, we focus on harvesting, or mining, thoracic disease localizations from chest X-ray images. Harvesting such localizations from existing datasets allows for the creation of improved data sources for computer-aided diagnosis and retrospective analyses. We train a convolutional neural network (CNN) for image classification and propose an attention mining (AM) strategy to improve the model's sensitivity or saliency to disease patterns. The intuition of AM is that once the most salient disease area is blocked or hidden from the CNN model, it will pay attention to alternative image regions, while still attempting to make correct predictions. However, the model requires to be properly constrained during AM, otherwise, it may overfit to uncorrelated image parts and forget the valuable knowledge that it has learned from the original image classification task. To alleviate such side effects, we then design a knowledge preservation (KP) loss, which minimizes the discrepancy between responses for X-ray images from the original and the updated networks. Furthermore, we modify the CNN model to include multi-scale aggregation (MSA), improving its localization ability on small-scale disease findings, \eg{}, lung nodules. We experimentally validate our method on the publicly-available ChestX-ray14 dataset, outperforming a class activation map (CAM)-based approach, and demonstrating the value of our novel framework for mining disease locations.
\end{abstract} 

\section{Introduction}
Automatic analysis of chest X-rays is critical for diagnosis and treatment planning of thoracic diseases. Recently, several methods applying deep learning for automatic chest X-ray analysis~\cite{rajpurkar_2017_chexnet,li_2017_thoracic,wang_2017_chestx,yao_2017_chestxray,pesce_2017_nodule} have been proposed. In particular, much work has focused on the ChestX-ray14 dataset~\cite{wang_2017_chestx}, which is an unprecedentedly large-scale and rich dataset but only provides image-level labels for the far majority of the samples. On the other hand, harvesting abnormality locations in this dataset is an important goal, as that provides an even richer source of data for training computer-aided diagnosis system and{\small /}or performing retrospective data analyses. Harvesting disease locations can be conducted through a weakly supervised image classification approach~\cite{wang_2017_chestx}; or, in our case we reformulate it as a label supervised \emph{pattern-mining problem}, to gain higher localization accuracy. Toward this end, we propose an integrated and novel framework that combines attention mining, knowledge preservation, and multi-scale aggregation that improves upon current efforts to accurately localize disease patterns. 

\par \qquad Recent work on chest X-rays have focused on both classification and localization. Along with the ChestX-ray14 dataset, Wang \etal{}~\cite{wang_2017_chestx} also propose a class activation map (CAM)-based~\cite{zhou_2016_cam} approach using convolutional neural network (CNNs) to perform weakly supervised disease localization. To improve image classification accuracy, Rajpurkar \etal{}~\cite{rajpurkar_2017_chexnet} introduce an ultra-deep CNN architecture while Yao \etal{}~\cite{yao_2017_chestxray} design a new learning objective that exploits dependencies among image labels. Other work investigate methods to automatically generate X-ray reports~\cite{Jing2017,Wang2018}. The proposed framework is a complementary or orthogonal development from the above advances~\cite{rajpurkar_2017_chexnet,yao_2017_chestxray} since we mine ``free'' disease locations in the form of bounding boxes given image-level labels. It also can further benefit downstream applications like~\cite{Jing2017} and~\cite{Wang2018}. 

\par \qquad  In terms of related work, our attention mining (AM) approach is closely related to an adversarial erasing scheme proposed in~\cite{Wei2017} that forces the network to discover other salient image regions by erasing the most representative area of the object class in question. In a similar spirit, we propose AM to locate multiple suspicious disease regions inside a chest X-ray. However, different from~\cite{Wei2017}, AM drops out corresponding pixels in the activation maps so as to leave the original X-ray images unchanged. More importantly, AM is designed to seamlessly couple with multi-label classification, where activation maps are required to be blocked in a class-wise manner. Next, to alleviate the side effects caused by dropping out activation maps, we exploit methods to prevent the network from forgetting its originally learned knowledge on recognizing and localizing disease patterns. Distilling a network's knowledge is proposed in~\cite{Hinton2015} to transfer the learned parameters from multiple models to a new, typically smaller sized, model. A similar technique is used in~\cite{kon_2017_forgetting} to regularize the CNN model for incremental learning with new image categories, keeping the network's output of old image categories mostly unchanged. In our method, we minimize the {\small $\ell_2$}-distance of output logits between the original and updated networks to achieve knowledge preservation (KP). Distinct from~\cite{Hinton2015} and~\cite{kon_2017_forgetting}, we use the logits not only from the last output layer but also the intermediate network layers, in order to introduce stronger regularizations. Finally, we propose a multi-scale aggregation (MSA) because we notice that the localization accuracy of lung nodules in~\cite{wang_2017_chestx} is not as good as the other disease findings, which we believe results from the coarse resolution of the attention maps, \ie{}, CAMs. Inspired by recent work~\cite{h_noh_iccv15_deconvnet,yu_2015_msa} we modify the CNN to generate attention maps with doubled resolution, improving the detection performance of small-scale targets. 

\section{Methods} \label{section:method}
\input{figures/framework}
Our framework is visually depicted in Fig.~\ref{fig:overview}. 

\subsection{Disease Pattern Localization with Attention Mining} \label{sec:am}
Starting from the output of CNN's last convolutional layer, we denote the feature map as {\small $X \in R^{N\times W\times H\times D}$}, where {\small $N$}, {\small $W$}, {\small $H$}, and {\small $D$} are the mini-batch size, width, height, and feature map channel, respectively. We then split the classification layer of the CNN into {\small $C$} branches because feature map erasure is required to be class specific. For now, we assume a binary erasure mask is available, which is defined as {\small $M^c \in R^{N\times W\times H\times 1}$}, where {\small $c \in \mathcal{C}=\{1,\ldots,C\}$} is the index of a specific disease type (see Section~\ref{exp:loc} for details to generate {\small $M^c$}). Zeroed regions in {\small $M^c$} mark spatial regions to drop out of {\small $X$}. For the {\small $c^{th}$} disease, {\small $M^c$} first replicated across its {\small $4^{th}$} dimension {\small $D$} times as {\small $\hat{M}^c$}, and then the erased feature map is, 
\begin{equation}  \label{eqn:erase_mask}
\hat{X}^c = X \odot \hat{M}^c,
\end{equation} 
where {\small $\odot$} is element-wise multiplication. The new feature map {\small $\hat{X}^c$} is then fed into the {\small $c^{th}$} network branch for binary classification, with the loss defined as,
\begin{equation}
L^c = \frac{1}{N}\left[h\left(\sigma\left((w^c)^Tg(\hat{X}^c)\right), y^c\right)\right], 
\end{equation}
where {\small $g(\cdot)$} is global average pooling (GAP)~\cite{zhou_2016_cam} over the {\small $W$} and {\small $H$} dimensions, {\small $w^c \in R^{D\times 1}$} is the network parameter of the {\small $c^{th}$} branch, {\small $\sigma(\cdot)$} is the sigmoid activation function, {\small $y^c \in \{0,1\}^{N}$} are the labels of class {\small $c$} in a mini-batch, and {\small $h(\cdot)$} is the cross entropy loss function. Thus, the total classification loss is defined as, 
\begin{equation}
L_{cls} = \frac{1}{C}\sum_{c \in \mathcal{C}} L^c .
\end{equation}
\par \qquad While AM can help localize pathologies, the CNN model may overfit to spurious regions after erasure, causing the model to classify an X-ray by remembering its specific image part rather than actual disease patterns. We address this with a knowledge preservation (KP) method described below. 

\subsection{Incremental Learning with Knowledge Preservation} \label{sec:kp}
We explore two methods of KP. Given a mini-batch of {\small $N$} images, a straightforward way to preserve the learned knowledge is to use only the first {\small $n$} images for AM and leave the later {\small $N-n$} untouched. If the ratio {\small $n/N$} is set to be small enough (\eg, 0.125 in our implementation), the CNN's updates can possibly be alleviated from overfitting to uncorrelated image parts. We refer to this vanilla implementation of knowledge preservation as KP-Vanilla. 

We investigate a stronger regularizer for KP by constraining the outputs of intermediate network layers. Our main idea is to make the CNN's activation to the later {\small $N-n$} images be mostly unchanged. Formally, we denote the original network before AM updates as {\small $\mathcal{N}_A$} and the updated model as {\small $\mathcal{N}_B$}. Initially, {\small $\mathcal{N}_A$} and {\small $\mathcal{N}_B$} are identical to each other, but {\small $\mathcal{N}_B$} is gradually altered as it learns to classify the blocked feature maps during AM. Considering outputs from the {\small $k^{th}$} layer of {\small $\mathcal{N}_A$} and {\small $\mathcal{N}_B$} as {\small $X_{A(k)}, X_{B(k)} \in R^{(N-n)\times W\times H\times C}$} for the later {\small $N-n$} images, we define the distance between {\small $X_{A(k)}$} and {\small $X_{B(k)}$} as the {\small $\ell_2$}-distance between their GAP features as,
\begin{equation}
L^k = \frac{1}{N-n} \| g(X_{A(k)}) - g(X_{B(k)})\|_2 .
\end{equation}
When multiple network layers are chosen, the total loss from KP is,
\begin{equation}
L_{KP} = \frac{1}{|\mathcal{K}|}~\sum_{k \in \mathcal{K}} L^k, 
\end{equation}
where {\small $\mathcal{K}$} is the indices set of the selected layers, and {\small $|\mathcal{K}|$} is its cardinality. Finally, the objective for {\small $\mathcal{N}_B$} training is a weighted combination of {\small $L_{cls}$} and {\small $L_{KP}$}, 
\begin{equation} \label{eq:loss}
L = L_{cls} + \lambda L_{KP}, 
\end{equation}
where {\small $\lambda$} balances the classification and KP loss. Empirically we find the model updates properly when the value of {\small $\lambda L_{KP}$} is roughly a half of {\small $L_{cls}$}, \ie, {\small $\lambda=0.5$}.

\subsection{Multi-Scale Aggregation} \label{method:msa}
Our final contribution uses multi-scale aggregation (MSA) to improve the performance of locating small-scale objects, \eg, lung nodules. Taking ResNet-50~\cite{he_2016_resnet} as the backbone network, we implement MSA using the outputs of the last two bottlenecks, and refer to the modified network as ResNet-MSA. Given the output of the last bottleneck, denoted as {\small $X_{k} \in R^{N\times W/2\times H/2\times 2048}$}, we feed it into a {\small $1\times 1$} convolutional layer to reduce its channel dimension to {\small $512$} and also upsample its width and height by {\small $2$} using bilinear interpolation. The resulting feature map is denoted as {\small $\bar{X}_k \in R^{N\times W\times H\times 512}$}. Similarly, the output of the penultimate bottleneck, {\small $X_{k-1} \in R^{N\times W\times H\times 1024}$}, is fed into another {\small $1\times 1$} convolutional layer to lower its channel dimension to {\small $256$}, producing {\small $\bar{X}_{k-1} \in R^{N\times W\times H\times 256}$}. Finally, we concatenate them to produce an aggregated feature map {\small $X = [\bar{X}_k,\bar{X}_{k-1}]$} for AM. However, MSA is not restricted to bilinear upsampling, as deconvolution~\cite{h_noh_iccv15_deconvnet} can also be used for upsampling, where we use {\small $3\times 3$} convolutions. However, as our experiments will demonstrate, the improvements are marginal, leaving bilinear as an efficient option. On the other hand, the channel dimensions of {\small $X_k$} and {\small $X_{k-1}$} are largely reduced in order to fit the models into limited GPU memory.  

\section{Experimental Results and Analysis} 
The proposed method is evaluated on the ChestX-ray14 dataset~\cite{wang_2017_chestx}, which contains {\small $51,709$} and {\small $60,412$} X-ray images of subjects with thoracic and no diseases, respectively.  {\small $880$} images are marked with bounding boxes (bboxs) corresponding to {\small $984$} disease patterns of 8 types, \ie, atelectasis (AT), cardiomegaly (CM), pleural effusion (PE), infiltration (Infiltrat.), mass, nodule, pneumonia (PNA), and pneumothorax (PTx). We first use the same data split as~\cite{wang_2017_chestx} to train base models, \ie, ResNet-50, and ResNet-MSA. Later during AM, the {\small $880$} bbox images are then incorporated into the training set to further fine-tune models. We notice that the AM strategy is originally designed to mine disease locations in training images. However, for the purpose of conducting quantitative analysis, we use the bbox images during AM, but only using image labels for training, while leaving the bboxs aside to evaluate localization results.

\par \qquad For ease of comparison, we use the same evaluation metrics as~\cite{wang_2017_chestx}. Given the ground truth and the localized bboxs of a disease, its localization accuracy (Acc.) and average false positive (AFP) is calculated by comparing the intersection over union (IoU) ratios with a predefined threshold, \ie, T(IoU). Finally, all of our deep learning implementations are built upon Tensorflow~\cite{tensorflow2015-whitepaper} and Tensorpack~\cite{wu2016tensorpack}.  

\subsection{Multiple Scale Aggregation} \label{exp:msa}
We first test the impact of MSA prior to the application of AM and KP,  implementing the bilinear interpolation and deconvolution variants. We also test two different input image resolutions: {\small $1024\times 1024$} and {\small $512\times 512$}, where the latter is downsampled from the original images using bilinear interpolation. Before applying MSA, we fine-tune the base network ResNet-50 with a learning rate of {\small $0.1$} for {\small $50$} epochs.  Mini-batch sizes for {\small $1024$} and {\small $512$} inputs are {\small $32$} and {\small $64$}, respectively. Then, to initialize MSA, we fix the network parameters below MSA and tune the other layers for 10 epochs. Finally, we have the whole ResNet-MSA updated end-to-end until the validation loss plot plateaus. Since we mainly focus on investigating AM and KP, no further modification has been taken for the network architecture, and thus the ResNet-MSA achieves similar classification performance as reported in~\cite{wang_2017_chestx} (see supplementary materials for details).

\par \qquad The results of different MSA setups are reported in Table~\ref{tab:msa}, where the ``baseline'' refers to the original ResNet-50, the ``bilinear'' and ``deconv.'' refer to ResNet-MSA with bilinear upsampling, and deconvolution operation, respectively. Prefixes denote the input resolution. As can be seen,  the {\small $512$} variants perform better than their {\small $1024$} counterparts. This is likely because the receptive field size of the MSA layers with ``{\small $1024$}-'' input is too small to capture sufficient contextual information. Note that for the ``{\small $512$}-'' input, the two MSA configurations outperform the baseline by a large margin for the infiltration, mass, and nodule categories. This is supporting our design intuition that MSA can help locate small-scale disease patterns more accurately. Because of the efficiency of bilinear upsampling, we select it to be MSA variant of choice, which will be further fine-tuned with AM and KP using Equation~\eqref{eq:loss}. \input{data/localization_msa}

\subsection{Disease Pattern Localization with AM and KP} \label{exp:loc}
In our implementation, we develop the attention mining (AM) basing on the class activation map (CAM) approach~\cite{zhou_2016_cam} that obtains class-specific heat maps. Specifically, the binary erasure mask, {\small $M^c$} is initialized to be all {\small $1$}, denoted as {\small $M^c_0$}. The AM procedure is then iteratively performed {\small $T$} times, and at time step {\small $t$}, the intermediate CAMs are generated as, 
\begin{equation} \label{eqn:heatmap_gen}
	H^{c}_{t}=(X \odot \hat{M}^c_{t-1})w^c ,
\end{equation}
where the inner product is executed across the channel dimension. These CAMs are then normalized to {\small$[0,1]$} and binarized with a threshold of {\small $0.5$}. {\small $M^c_{t}$} is then updated from {\small $M^c_{(t-1)}$}, except that pixel locations of the connected component that contains the global maximum of the binarized CAM are now set to {\small $0$}.

\par \qquad There are different options to generate a final heatmap aggregated from all {\small $T$} CAMs. We choose to have them averaged. However, when {\small $t>1$} regions have been erased, as per Equation~\eqref{eqn:heatmap_gen}. Thus, we fill in these regions from the corresponding un-erased regions from prior heatmaps. If we define the complement of the masks as {\small $\bar{M}^c_t = (1 - \hat{M}^c_t)$},  then the final heatmap {\small $H^c_{f}$} is calculated using
\begin{equation}
	H^c_{f} = \frac{1}{T} \sum_{t=1}^{T} \left[H^c_t + \sum_{t'=1}^{t-1}(H^c_{t' }\odot \bar{M}^c_{t'})\right]\textrm{.} 
\end{equation}
Empirically, we find {\small $T=3$} works best in our implementation. 

\textbf{Bbox Generation:} To convert CAMs into bboxs, we have $3$ bboxs generated from each $H^c_{f}$ by adjusting the intensity threshold. For image $i$, the bboxs are then ranked as $\{bbox^c_i(1), bbox^c_i(2), bbox^c_i(3)\}$ in descending order based on the mean $H^c_{f}$ intensity values inside the bbox areas. These are then arranged into an aggregated list across all test images from the {\small $c^{th}$} category: \\
\centerline{$\mathcal{B} = \{bbox^c_1(1),\ldots,bbox^c_N(1), bbox^c_1(2),\ldots,bbox^c_N(2), bbox^c_1(3),\ldots,bbox^c_N(3)\}$.} \\
Thereafter, these bboxs are sequentially selected from {\small $\mathcal{B}$} to calculate {\small $Acc.$} until the AFP reaches its upper bound, which is the corresponding AFP value reported in~\cite{wang_2017_chestx}. Here, we choose to generate {\small $3$} bboxs from each image as it is large enough to cover the corrected locations, while an even larger alternate will greatly increase the AFP value. However, in some cases, $H^c_f$ would just allow to generate fewer than {\small $3$} bboxs,  for instance, see Fig.~\ref{fig:show-am}(a).

\par \qquad Since Wang \etal{}~\cite{wang_2017_chestx} were not tackling the disease localization in the way as data-mining, direct comparison to their results is not appropriate as they incorporated the bbox images in their test set. Instead, we use our method prior to the application of AM steps as the baseline, which is, for all intents and purposes, Wang \etal{}'s approach~\cite{wang_2017_chestx} applied to the \emph{data-mining problem}. It is presented as the ``baseline'' method in Table~\ref{tab:cmpr}. More specifically, it is set up as the {\small $1^{st}$} time step of AM with KP-Vanilla and then fine-tuned until it is converged on the bbox images. As shown in Table~\ref{tab:cmpr}, our method reports systematic and consistent quantitative performance improvements over the ``baseline'' method, except slightly degrades on the category of CM, demonstrating the impact of our AM and KP enhancements. Meanwhile, comparing with the results in~\cite{wang_2017_chestx}, our method achieves significant improvements by using no extra manual annotations. More Importantly, the results in Table~\ref{tab:cmpr} indicate our method would also be effective when implemented to mine disease locations in the training images.

\input{data/localization_IoU} \input{figures/show_AM}

\par \qquad Figure~\ref{fig:show-am} depicts three atelectasis cases visualizing the AM process. As can be seen, AM improves upon the baseline results, {\small $H^c_1$}, by discovering new regions after erasing that correlate with the disease patterns. More qualitative results can be found in the supplementary materials. 

\textbf{Ablation Study:} We further investigate the impact of AM and KP. First, we compare the three steps in AM. Since the erasure map {\small $M^c_0$} is initialized with all 1s, then the time step {\small $t=1$} is treated as the baseline. Table~\ref{tab:ablation} shows the localization results at two IoU thresholds, $0.1$ and $0.3$. As can be seen, significant improvements of AM are observed in the AT, CM, mass, PNA, and PTx disease patterns. Next, we compare the KP, KP-Vanilla and an implementation without KP, where the ResNet-MSA is tuned in AM using only the bbox images. In particular, we set {\small $\mathcal{K}$} by using the outputs of the $3^{rd}$ and $4^{th}$ bottleneck, the MSA, and the classification layers. As Table~\ref{tab:ablation} presents, KP performs better than KP-Vanilla in the AT, PE, mass, nodule, PNA and PTx categories.
\input{data/localization_ablation}

\section{Conclusion}
We present a novel localization data-mining framework, combining AM, KP, and MSA. We introduce a powerful means to harvest disease locations from chest X-ray datasets. By showing improvements over a standard CAM-based approach, our method can mine localization knowledge in existing large-scale datasets, potentially allowing for the training of improved computer-aided diagnosis tools or more powerful retrospective analyses. Future work includes improving the MSA, possibly by using the atrous convolution~\cite{yu_2015_msa}. Additionally, we find that when the activation map fails to localize disease in none of the AM steps, our method will not locate the correct image region as demonstrated in Figure~\ref{fig:show-am}(c). To address this issue, we may consider semi-supervised learning, like the use of bboxs in~\cite{li_2017_thoracic}, as a complementary means to discover those difficult cases.

\bibliographystyle{splncs}
\bibliography{egbib}

\newpage
\section*{Supplementary Material}
\vspace{1cm}

\begin{table}[h!]
	\centering
	\caption{AUCs of ROC curves for thoracic disease classification for ChestX-ray14~\cite{wang_2017_chestx}. All of the results are reported by using ResNet-50~\cite{he_2016_resnet} as the base classification network. We notice the result reported from~\cite{li_2017_thoracic} is given by the network model that trained without bounding box annotations using the combinatorial loss function proposed in~\cite{li_2017_thoracic}.}
	\scriptsize{
		\begin{tabular}{l c c c c c c c c c c c c c c}
			\toprule
			Method \qquad \qquad & \rotatebox[origin=l]{90}{Atelectasis} & \rotatebox[origin=l]{90}{Cardiomegaly} & \rotatebox[origin=l]{90}{Consolidation} &\rotatebox[origin=l]{90}{Edema} & \rotatebox[origin=l]{90}{Effusion} & \rotatebox[origin=l]{90}{Emphysema} & \rotatebox[origin=l]{90}{Fibrosis} & \rotatebox[origin=l]{90}{Hernia} & \rotatebox[origin=l]{90}{Infiltration} & \rotatebox[origin=l]{90}{Mass} & \rotatebox[origin=l]{90}{Nodule} & \rotatebox[origin=l]{90}{Pleural Thickening ~} & \rotatebox[origin=l]{90}{Pneumonia} & \rotatebox[origin=l]{90}{Pneumothorax} \\
			\midrule
			Li~\etal{}~\cite{li_2017_thoracic} 		& \textBF{0.78} & 0.85 			& \textBF{0.79} & \textBF{0.85} & \textBF{0.86} & 0.89 			& 0.76 			& 0.68 			& 0.66 			& \textBF{0.81} & 0.72 			& \textBF{0.75} & 0.66 			& \textBF{0.85} \\
			Wang~\etal{}~\cite{wang_2017_chestx}    & 0.72 			& 0.81 			& 0.71 			& 0.83 			& 0.78 			& 0.81 			& 0.77 			& \textBF{0.77} & 0.61 			& 0.71 			& 0.67 			& 0.71 			& 0.63 			& 0.81 \\
			Ours ResNet-50 							& 0.71 			& 0.81 			& 0.70 			& 0.82 			& 0.79 			& 0.89 			& 0.79 			& 0.49 			& 0.66 			& 0.73 			& 0.70 			& 0.74 			& 0.67 			& 0.81  \\
			Ours ResNet-MSA        					& 0.75 		    & \textBF{0.86} & 0.69 			& 0.83 			& 0.81 			& \textBF{0.91} & \textBF{0.80} & 0.53 			& \textBF{0.67} & 0.80 			& \textBF{0.76} & \textBF{0.75} & \textBF{0.70} 			& \textBF{0.85}  \\
			\bottomrule
	\end{tabular} }
\end{table}

\begin{table}[h!]
	\centering
	\caption{Localization results comparison: This table shows the same localization results as Table~2 in the main manuscript. Instead of displaying differences, this table shows the localization performance as ``Acc.-AFP'', where ``ref.'' and ``base.'' are presenting the referred method proposed by Wang \etal{}~\cite{wang_2017_chestx} and the baseline method presented in Sec. 3.2 in the manuscript, respectively. The final results of implmenting AM and KP are presented as ``ours''. The best performance, which has the highest $Acc.$ and lowest AFP values, is presented in bold.}
	\vspace{0mm}
	\label{tab:cmpr-sup}
	\scriptsize{
		\begin{tabular}{c c c c c c c c c c}
			\toprule
			T(IoU) & Method & AT & CM & Effusion & Infiltra. & Mass & Nodule & PNA & PTx\\
			\midrule
			\rowcolor{Gray}
			& ref. & 0.69-0.89 			& 0.94-0.59 		 & 0.66-0.83		  & 0.71-0.62		 & 0.40-0.67 			& 0.14-0.61 		 & 0.63-1.02 		  & 0.38-0.49 \\ 
			\rowcolor{Gray}
			0.1 & base.& 0.55-0.88 			& \textBF{0.99-0.08} & 0.57-0.82 		  & 0.47-0.61 		 & 0.36-0.65 			& 0.25-0.59 		 & \textBF{0.73-1.01} & 0.42-0.48 \\
			\rowcolor{Gray}
			& ours & 0.68-0.88 		 	& 0.97-0.18 		 & 0.65-0.82 		  & 0.52-0.61 		 & \textBF{0.56-0.65} 	& \textBF{0.46-0.59} & 0.65-1.01 		  & \textBF{0.43-0.48} \\ 
			
			
			& ref. & 0.47-0.98 			& 0.68-0.72 		 & 0.45-0.91 		  & 0.48-0.68 			 & 0.26-0.69 		  & 0.05-0.62 		   & 0.35-1.08 		    & 0.23-0.52 \\
			0.2 & base.& 0.36-0.97 			& \textBF{0.99-0.08} & 0.33-0.90 		  & 0.22-0.67 			 & 0.26-0.68 		  & 0.09-0.61 		   & 0.48-1.07 			& \textBF{0.36-0.50} \\
			& ours & \textBF{0.51-0.97} & 0.90-0.28 		 & \textBF{0.52-0.90} & 0.44-0.67 		   	 & \textBF{0.47-0.68} & \textBF{0.27-0.61} & \textBF{0.54-1.07} & 0.24-0.50 \\
			
			
			\rowcolor{Gray}
			& ref. & 0.24-1.40 			& 0.46-0.78 		 & 0.30-0.95 		  & 0.28-0.72 		   & 0.15-0.71 			& 0.04-0.62 		 & 0.17-1.11 		  & 0.13-0.53 \\
			\rowcolor{Gray}
			0.3 & base.& 0.22-1.38 			& \textBF{0.96-0.12} & 0.18-0.93 		  & 0.13-0.71 		   & 0.19-0.69 			& 0.04-0.61 		 & 0.31-1.09 		  & \textBF{0.21-0.52} \\
			\rowcolor{Gray}
			& ours & \textBF{0.33-1.38} & 0.85-0.35 		 & \textBF{0.34-0.93} & \textBF{0.28-0.71} & \textBF{0.33-0.69} & \textBF{0.11-0.61} & \textBF{0.39-1.09} & 0.16-0.52 \\
			
			
			& ref. & 0.09-1.08 			& 0.28-0.81 		 & 0.20-0.97 & 0.12-0.75 		  & 0.07-0.72 			& 0.01-0.62 		 & 0.07-1.12 		  & 0.07-0.54 \\
			0.4 & base.& 0.12-1.06 			& \textBF{0.92-0.15} & 0.05-0.95 & 0.02-0.73 		  & 0.12-0.71 			& \textBF{0.03-0.61} & 0.17-1.11 		  & \textBF{0.12-0.53} \\
			& ours & \textBF{0.23-1.06} & 0.73-0.47 		 & 0.18-0.95 & \textBF{0.20-0.73} & \textBF{0.18-0.71}  & \textBF{0.03-0.61} & \textBF{0.23-1.11} & 0.11-0.53 \\
			
			\rowcolor{Gray}
			& ref. & 0.05-1.09 			& 0.18-0.84 		 & 0.11-0.99 & 0.07-0.76 		  & 0.01-0.72 			& 0.01-0.62 		 & 0.03-1.13 		  & 0.03-0.55 \\
			\rowcolor{Gray}
			0.5 & base.& 0.06-1.07 			& \textBF{0.68-0.39} & 0.02-0.97 & 0.02-0.74 		  & 0.06-0.71 			& 0.01-0.61 		 & 0.06-1.12 		  & \textBF{0.08-0.53} \\
			\rowcolor{Gray}
			& ours & \textBF{0.11-1.07} & 0.60-0.60 		 & 0.10-0.97 & \textBF{0.12-0.74} & \textBF{0.07-0.71}  & \textBF{0.03-0.61} & \textBF{0.17-1.12} & \textBF{0.08-0.53} \\

			& ref. & 0.02-1.09 			& 0.08-0.85 		 & 0.05-1.00 		  & 0.02-0.76 		   & 0.00-0.72 			& 0.01-0.62 		 & 0.02-1.13 		  & 0.03-0.55 \\
			
			0.6 & base.& 0.01-1.08 			& \textBF{0.48-0.60} & 0.00-0.99 		  & 0.02-0.75 		   & 0.04-0.71 			& \textBF{0.01-0.61} & 0.03-1.12 		  & 0.06-0.53 \\
			
			& ours & \textBF{0.03-1.08} & 0.44-0.76 		 & \textBF{0.05-0.99} & \textBF{0.06-0.75} & \textBF{0.05-0.71} & \textBF{0.01-0.61} & \textBF{0.05-1.12} & \textBF{0.07-0.53} \\
			
			\rowcolor{Gray}
			& ref. & 0.01-1.10 		    & 0.03-0.86 		 & 0.02-1.01 & 0.00-0.77 		  & 0.00-0.72 		   & 0.00-0.62 			& 0.01-1.13 		 & 0.02-0.55 \\
			\rowcolor{Gray}
			0.7 & base.& 0.00-1.08 			& \textBF{0.18-0.84} & 0.00-0.99 & 0.01-0.75 		  & \textBF{0.01-0.71} & \textBF{0.00-0.61} & 0.01-1.12 		 & 0.01-0.53 \\
			\rowcolor{Gray}
			& ours & \textBF{0.01-1.08} & 0.17-0.84 		 & 0.01-0.99 & \textBF{0.02-0.75} & \textBF{0.01-0.71} & \textBF{0.00-0.61} & \textBF{0.02-1.12} & \textBF{0.02-0.53} \\ 
			\bottomrule
		\end{tabular}
	}
\end{table}


\begin{figure}[h!]
	\begin{subfigure}{0.32\textwidth}
		\includegraphics[width=\linewidth, page=5, trim={7cm 0cm 7cm 0cm}, clip]{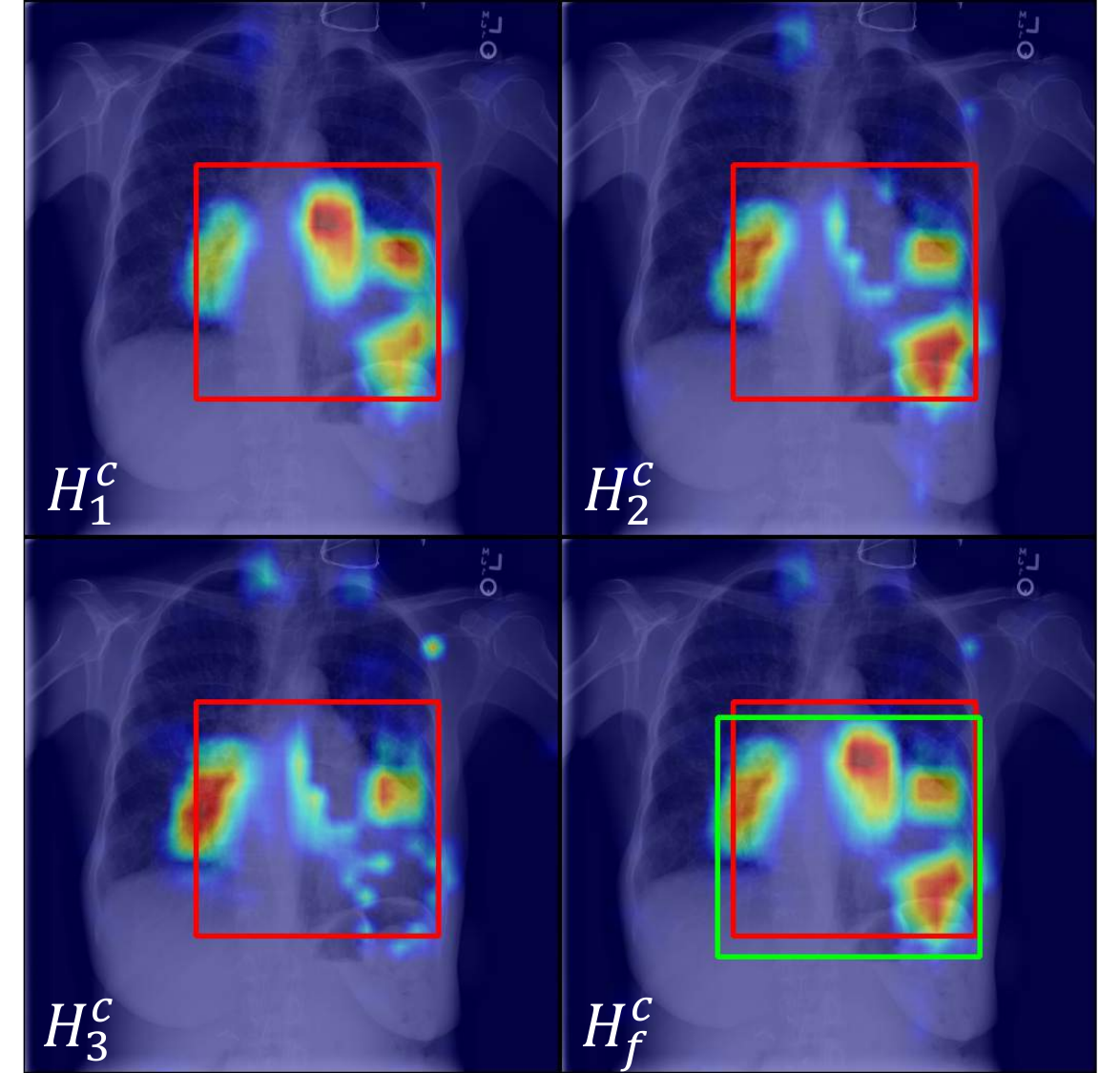}
	\end{subfigure}
	\begin{subfigure}{0.32\textwidth}
		\includegraphics[width=\linewidth, page=6, trim={7cm 0cm 7cm 0cm}, clip]{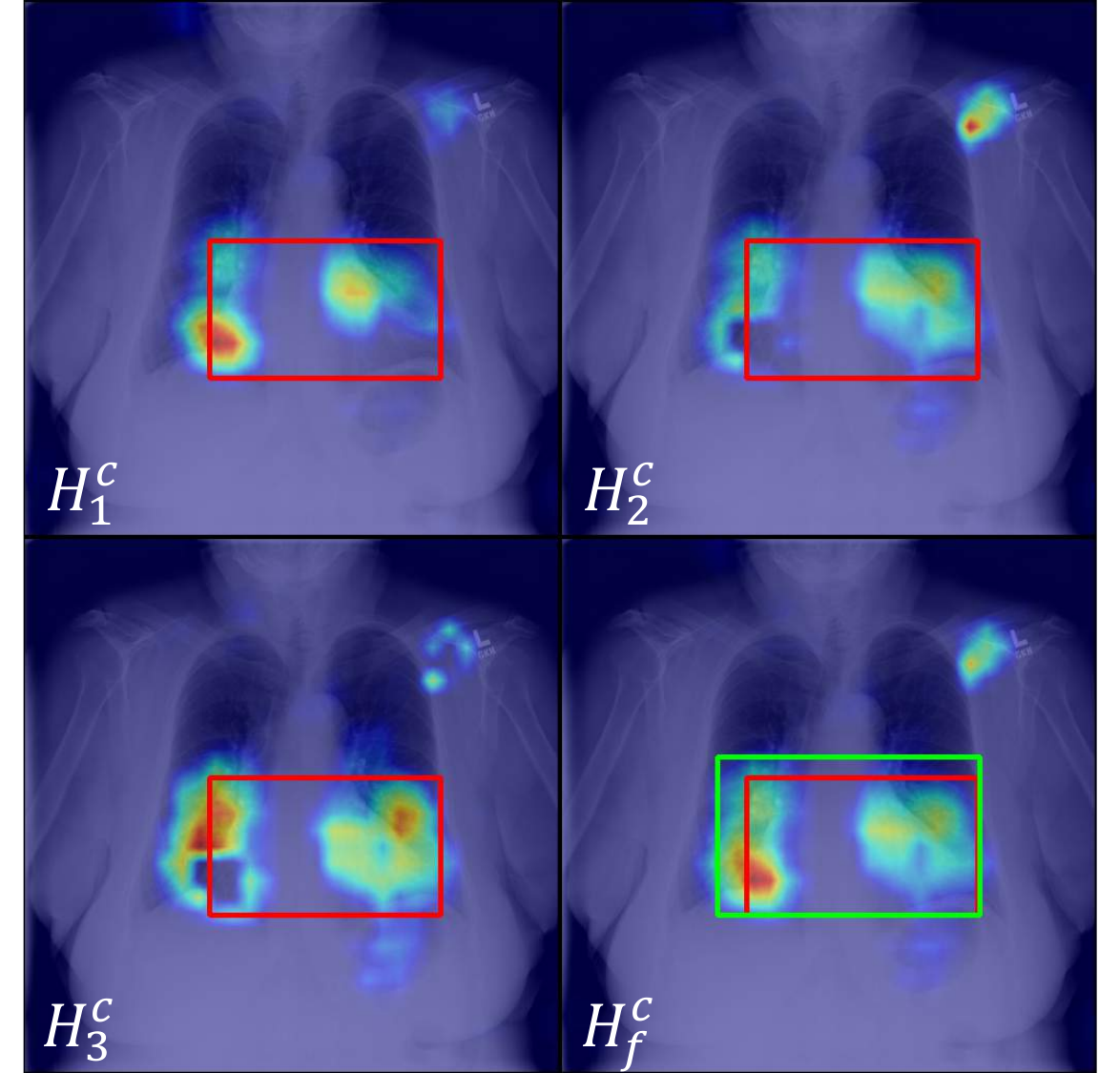}
	\end{subfigure}
	\begin{subfigure}{0.32\textwidth}
		\includegraphics[width=\linewidth, page=7, trim={7cm 0cm 7cm 0cm}, clip]{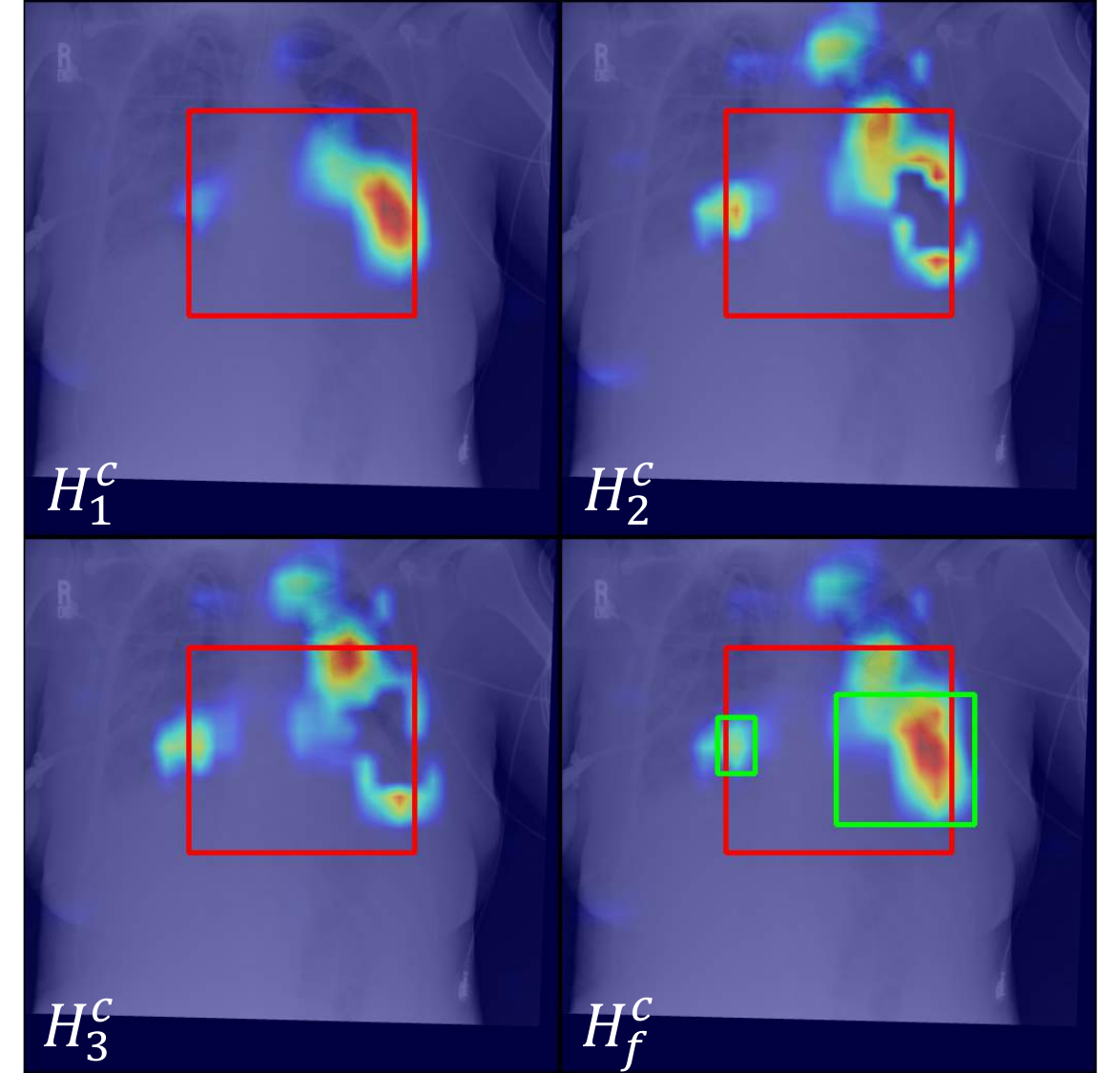}
	\end{subfigure}
	\vspace{0mm}
	\caption{Attetion mining on cardiomegaly. The automatically mined bounding boxes are generated from $H_f^c$ and displayed in green. The ground truth bounding boxes are given with red color. The same color scheme is also applied in the following figures.}
\end{figure}

\begin{figure}[h!]
	\begin{subfigure}{0.32\textwidth}
		\includegraphics[width=\linewidth, page=8, trim={7cm 0cm 7cm 0cm}, clip]{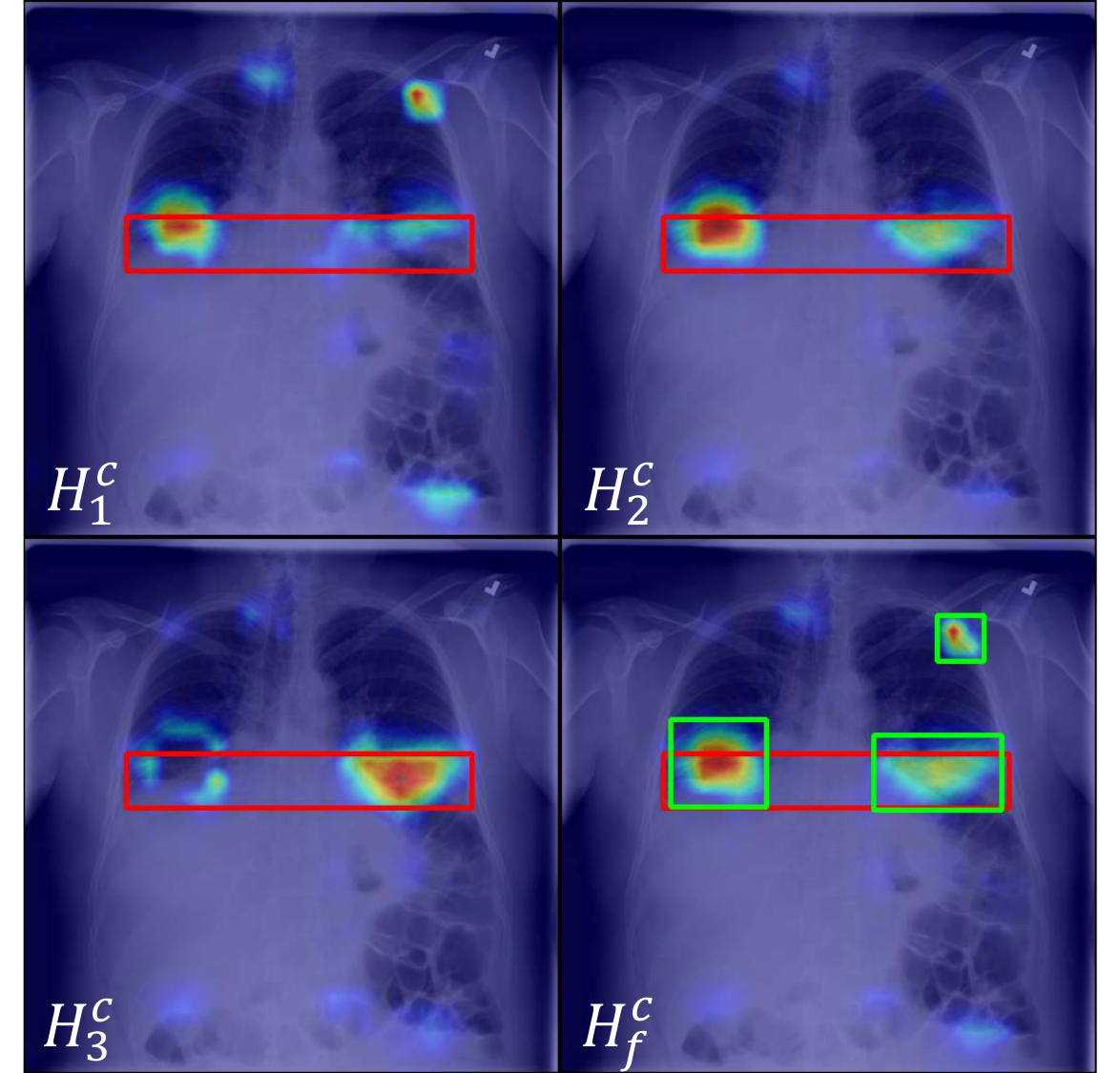}
	\end{subfigure}
	\begin{subfigure}{0.32\textwidth}
		\includegraphics[width=\linewidth, page=9, trim={7cm 0cm 7cm 0cm}, clip]{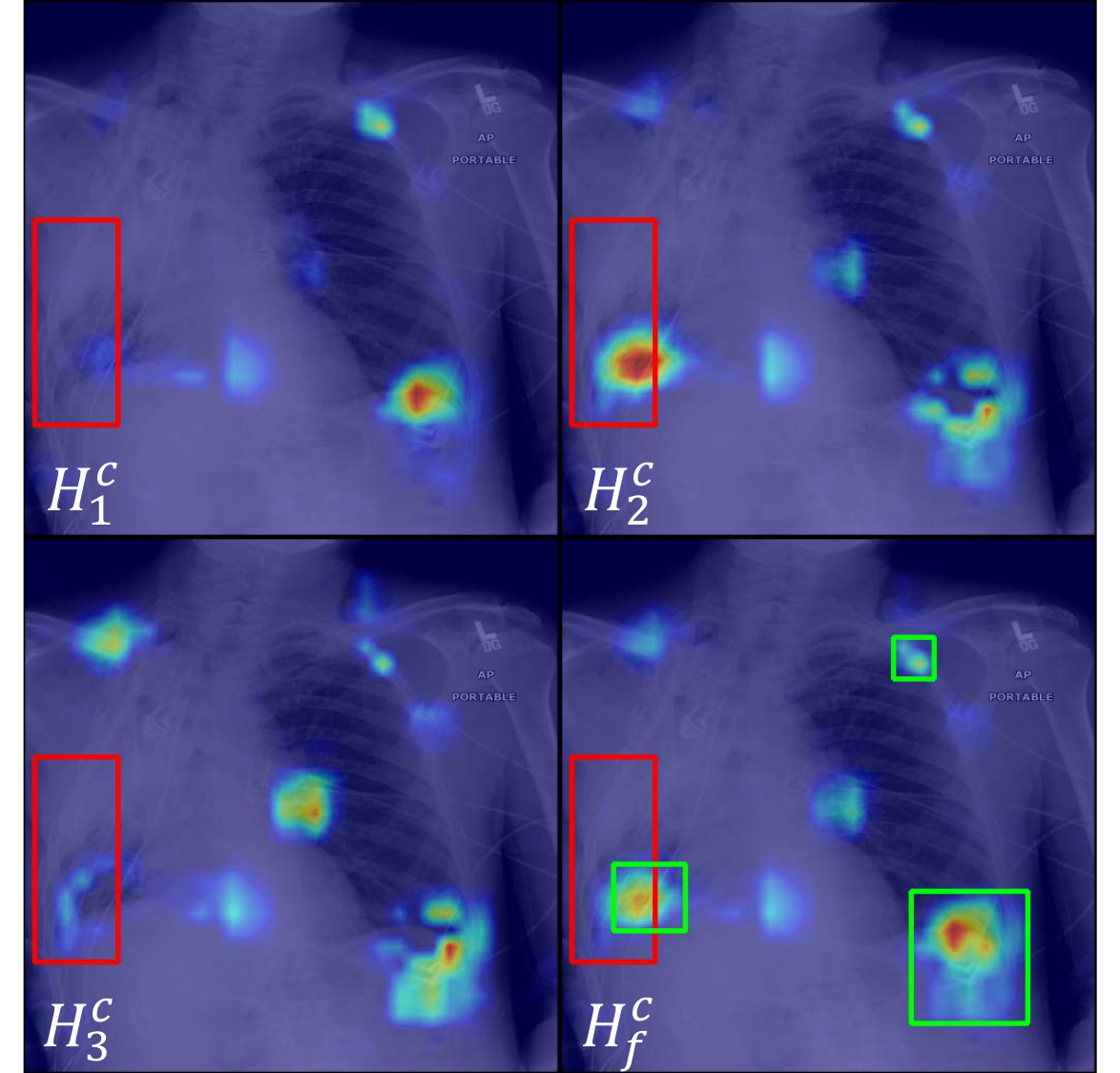}
	\end{subfigure}
	\begin{subfigure}{0.32\textwidth}
		\includegraphics[width=\linewidth, page=10, trim={7cm 0cm 7cm 0cm}, clip]{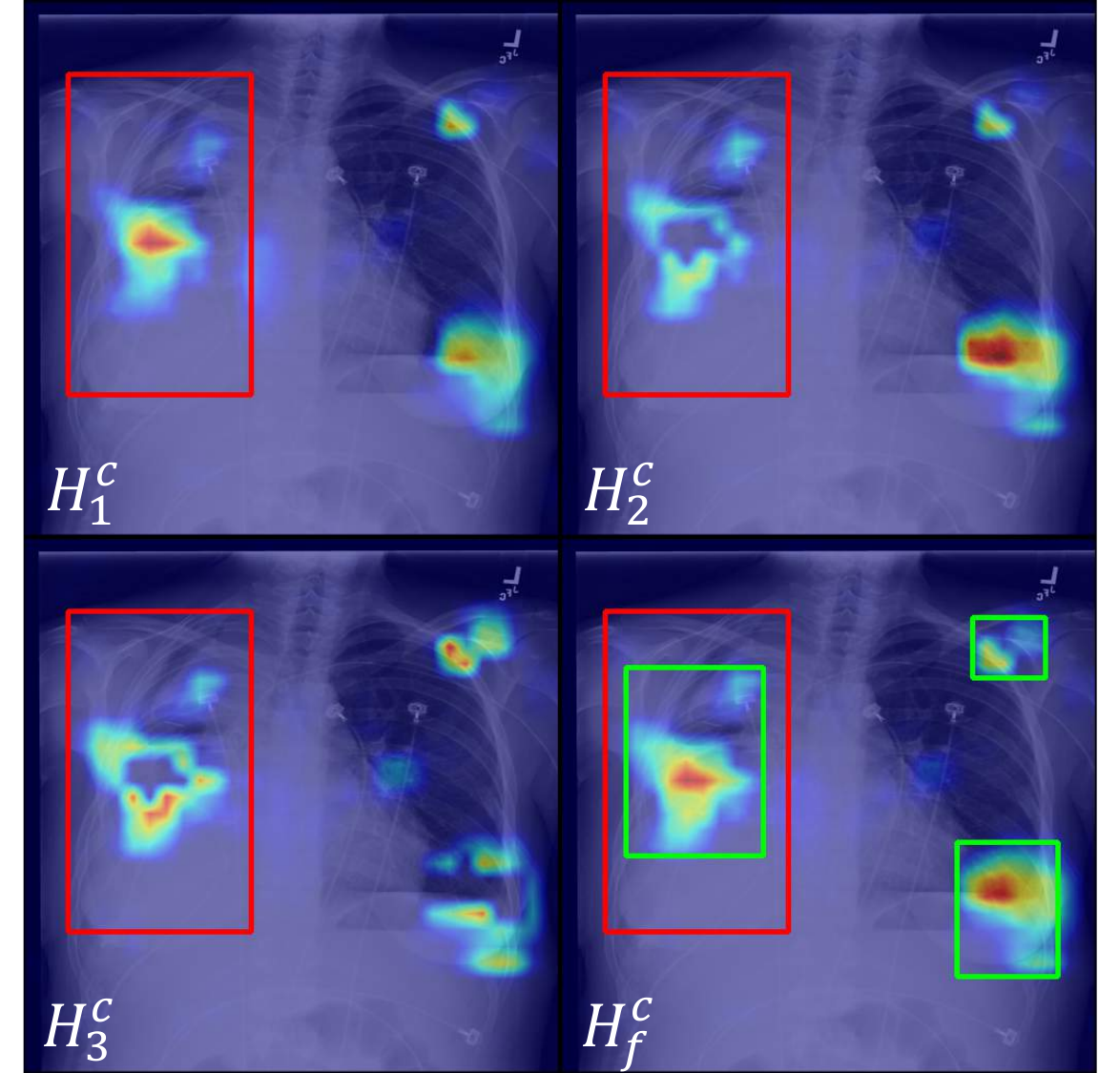}
	\end{subfigure}
	\vspace{0mm}
	\caption{Attetion mining on pleural effusion.}
\end{figure}

\begin{figure}[h!]
	\begin{subfigure}{0.32\textwidth}
		\includegraphics[width=\linewidth, page=11, trim={7cm 0cm 7cm 0cm}, clip]{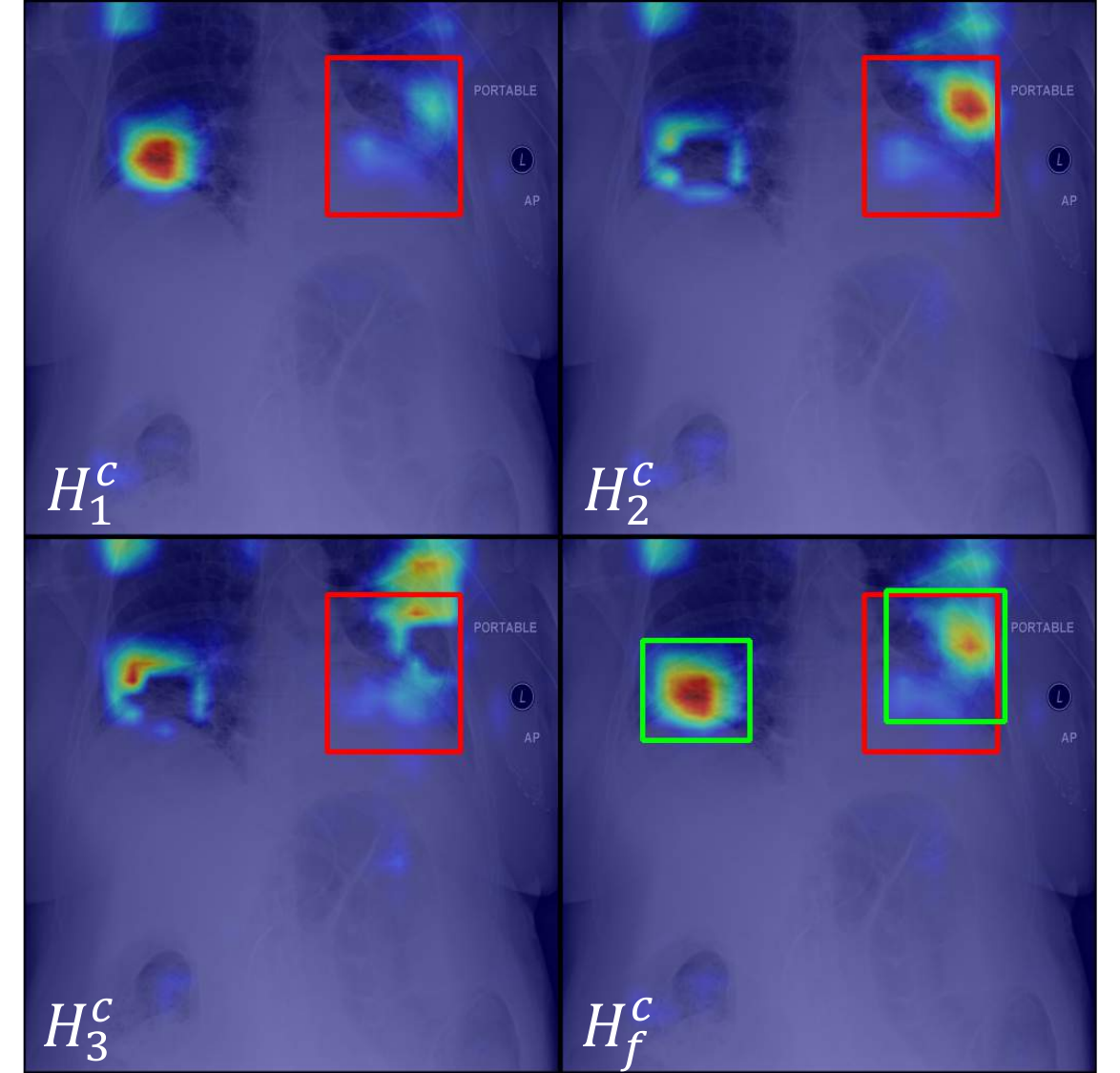}
	\end{subfigure}
	\begin{subfigure}{0.32\textwidth}
		\includegraphics[width=\linewidth, page=12, trim={7cm 0cm 7cm 0cm}, clip]{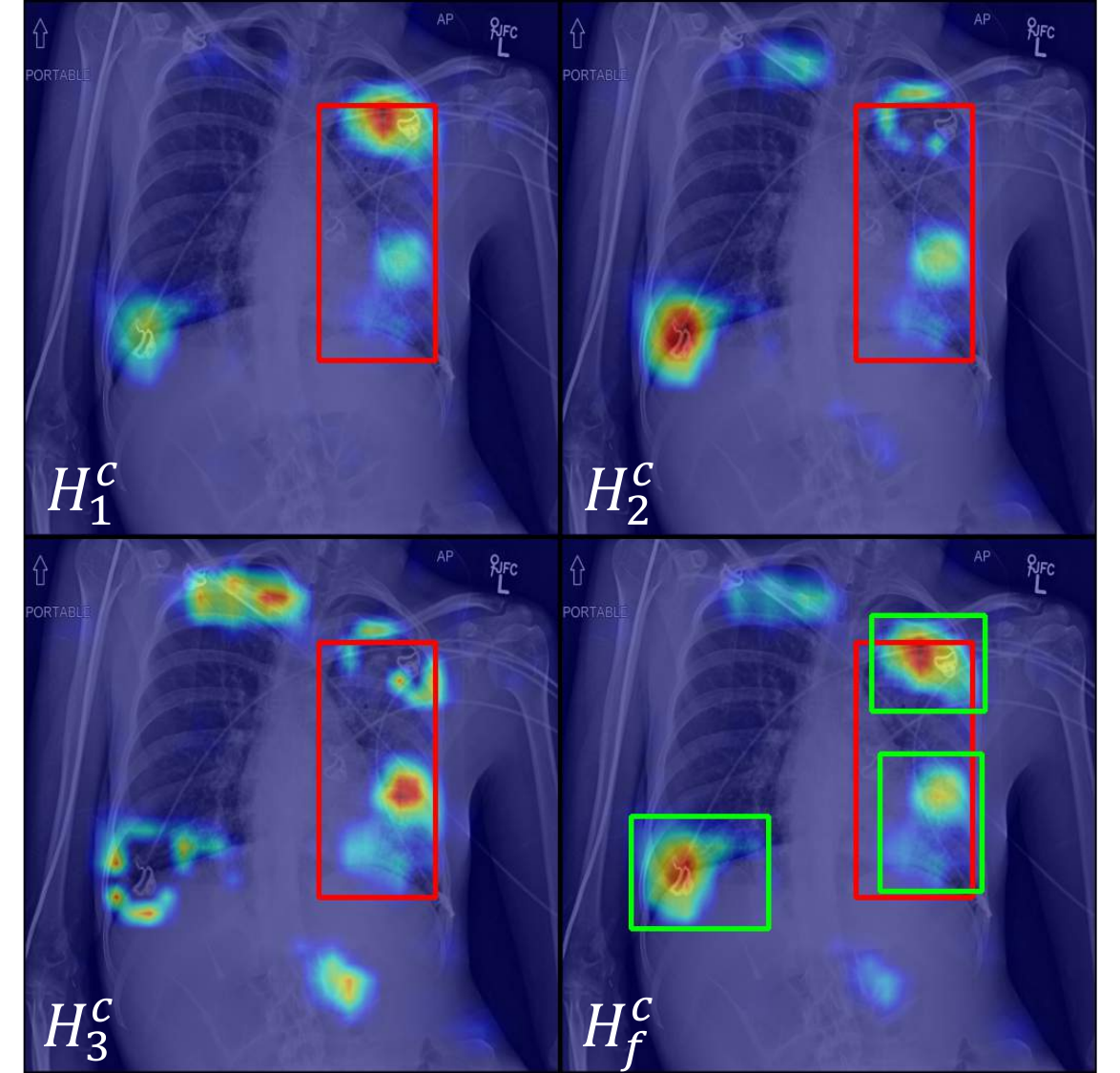}
	\end{subfigure}
	\begin{subfigure}{0.32\textwidth}
		\includegraphics[width=\linewidth, page=13, trim={7cm 0cm 7cm 0cm}, clip]{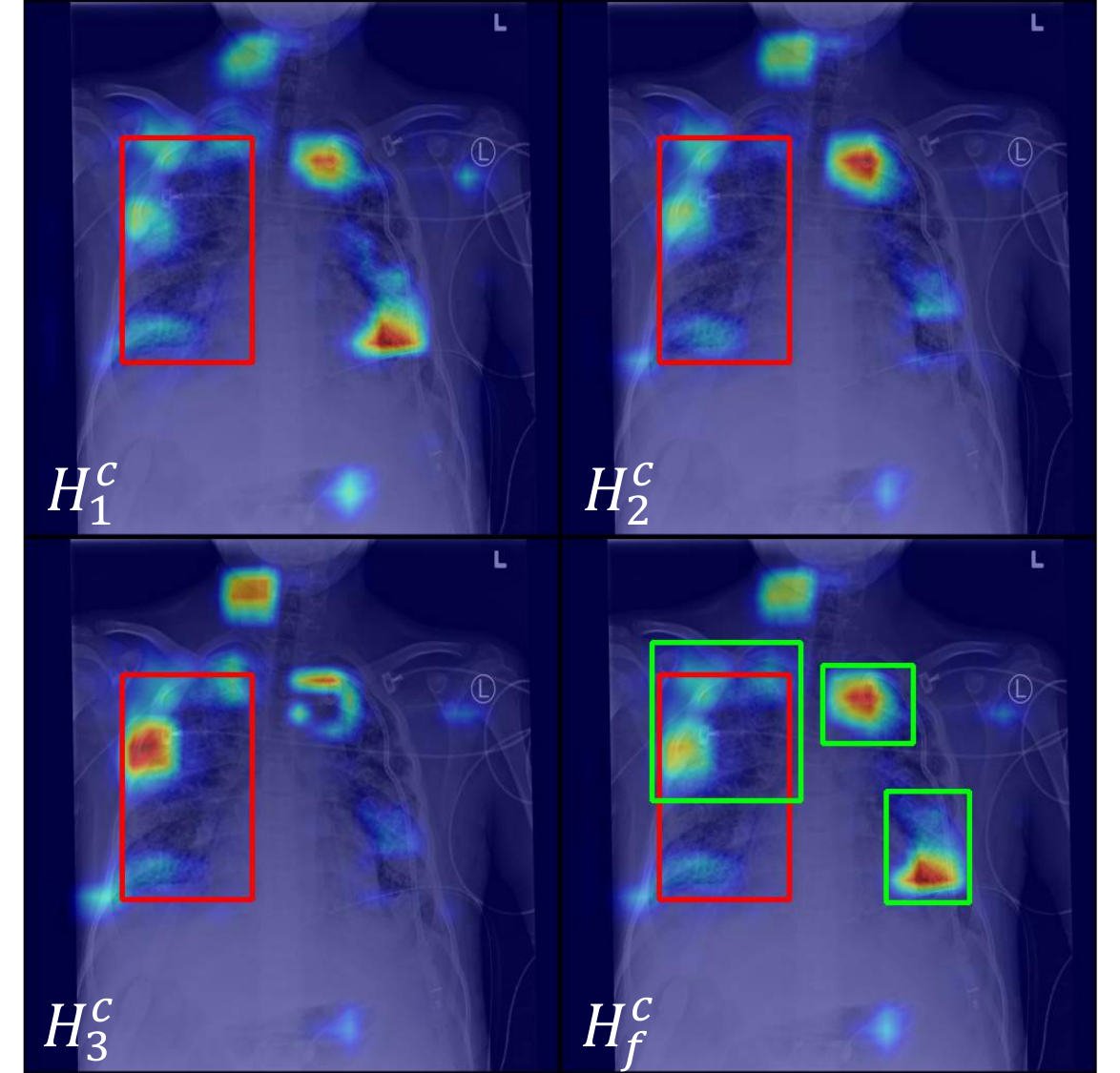}
	\end{subfigure}
	\vspace{0mm}
	\caption{Attetion mining on infiltration.}
\end{figure}

\begin{figure}[h!]
	\begin{subfigure}{0.32\textwidth}
		\includegraphics[width=\linewidth, page=14, trim={7cm 0cm 7cm 0cm}, clip]{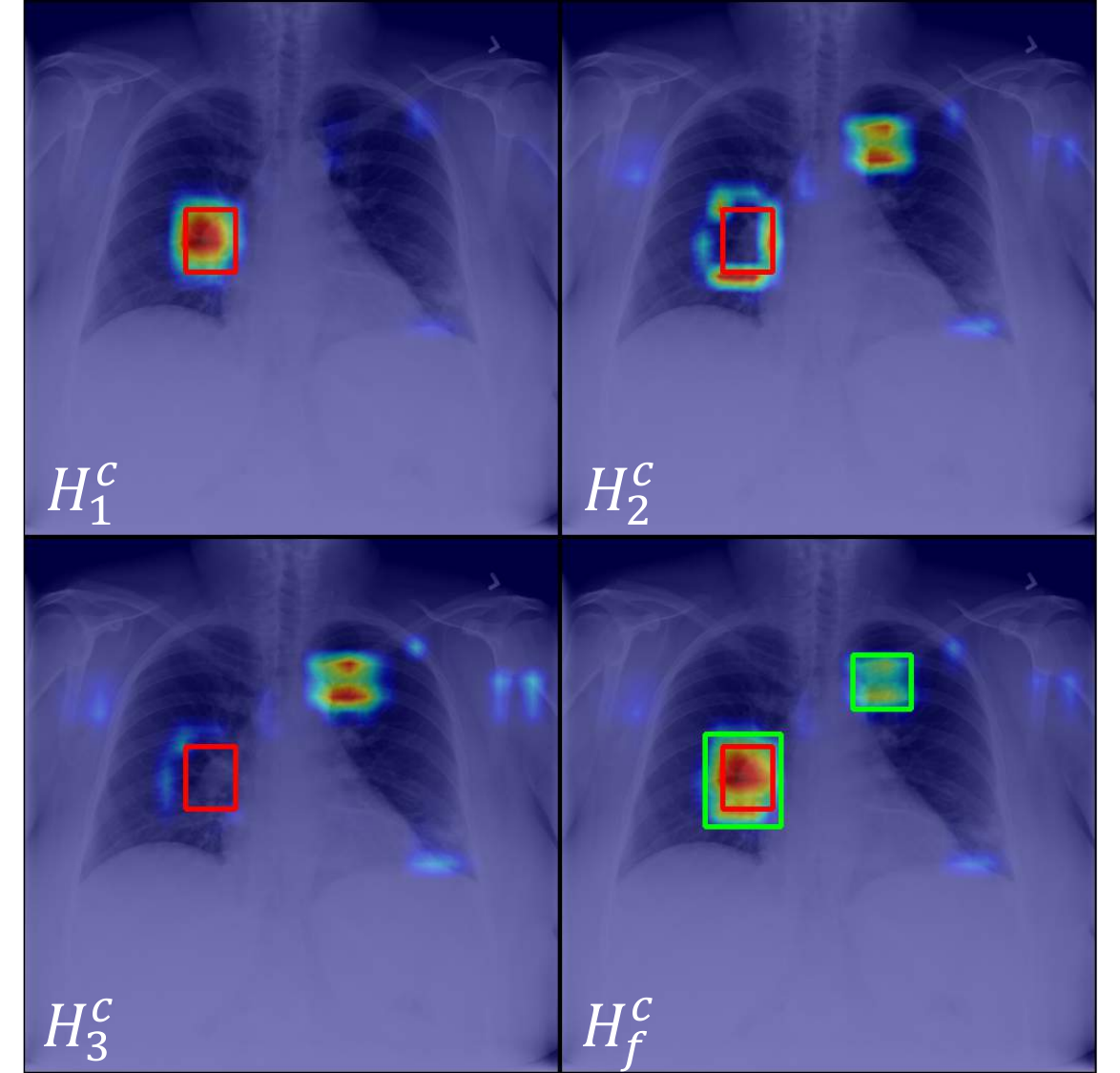}
	\end{subfigure}
	\begin{subfigure}{0.32\textwidth}
		\includegraphics[width=\linewidth, page=15, trim={7cm 0cm 7cm 0cm}, clip]{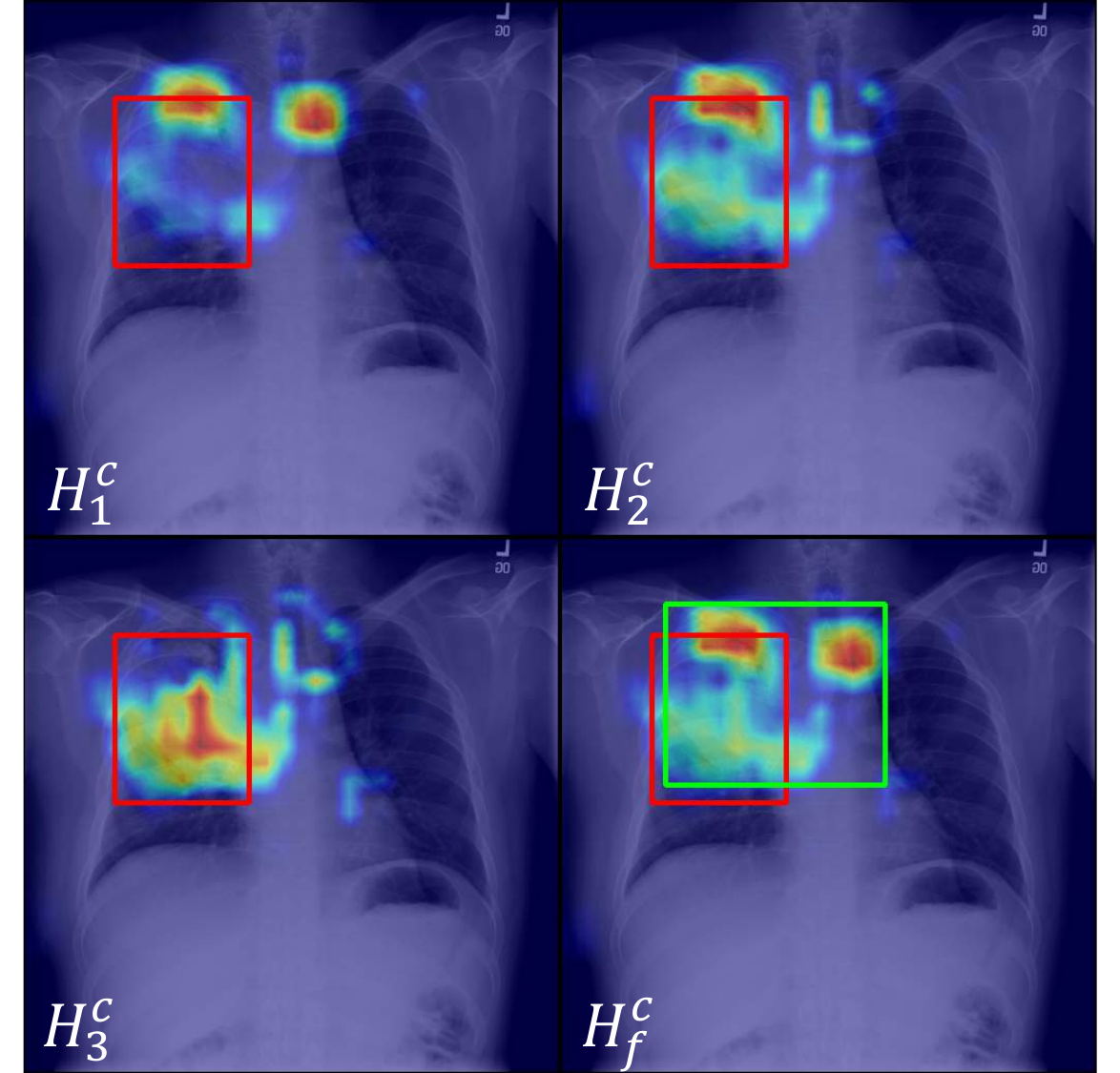}
	\end{subfigure}
	\begin{subfigure}{0.32\textwidth}
		\includegraphics[width=\linewidth, page=16, trim={7cm 0cm 7cm 0cm}, clip]{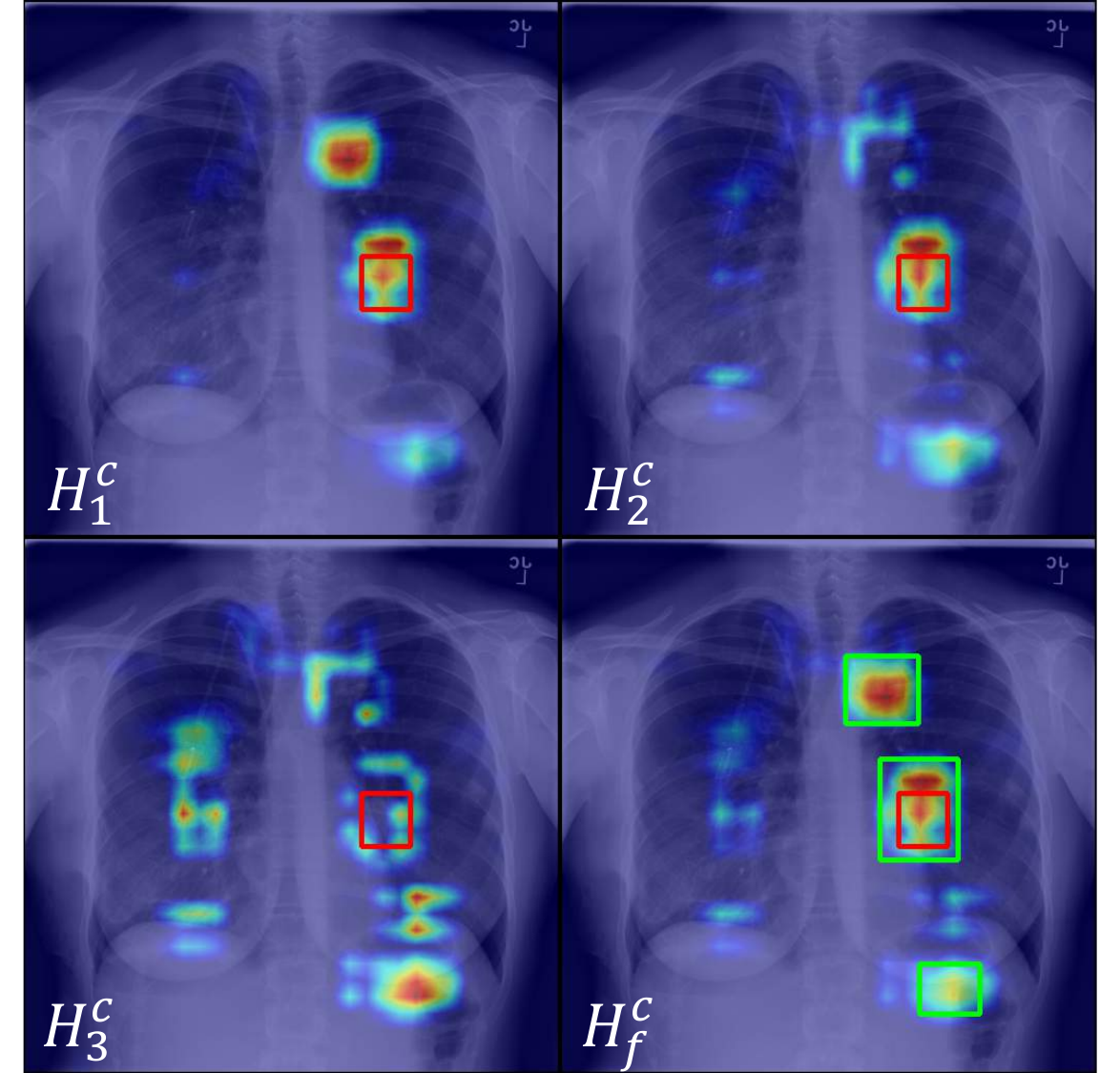}
	\end{subfigure}
	\vspace{0mm}
	\caption{Attetion ming on mass.}
\end{figure}

\begin{figure}[h!]
	\begin{subfigure}{0.32\textwidth}
		\includegraphics[width=\linewidth, page=17, trim={7cm 0cm 7cm 0cm}, clip]{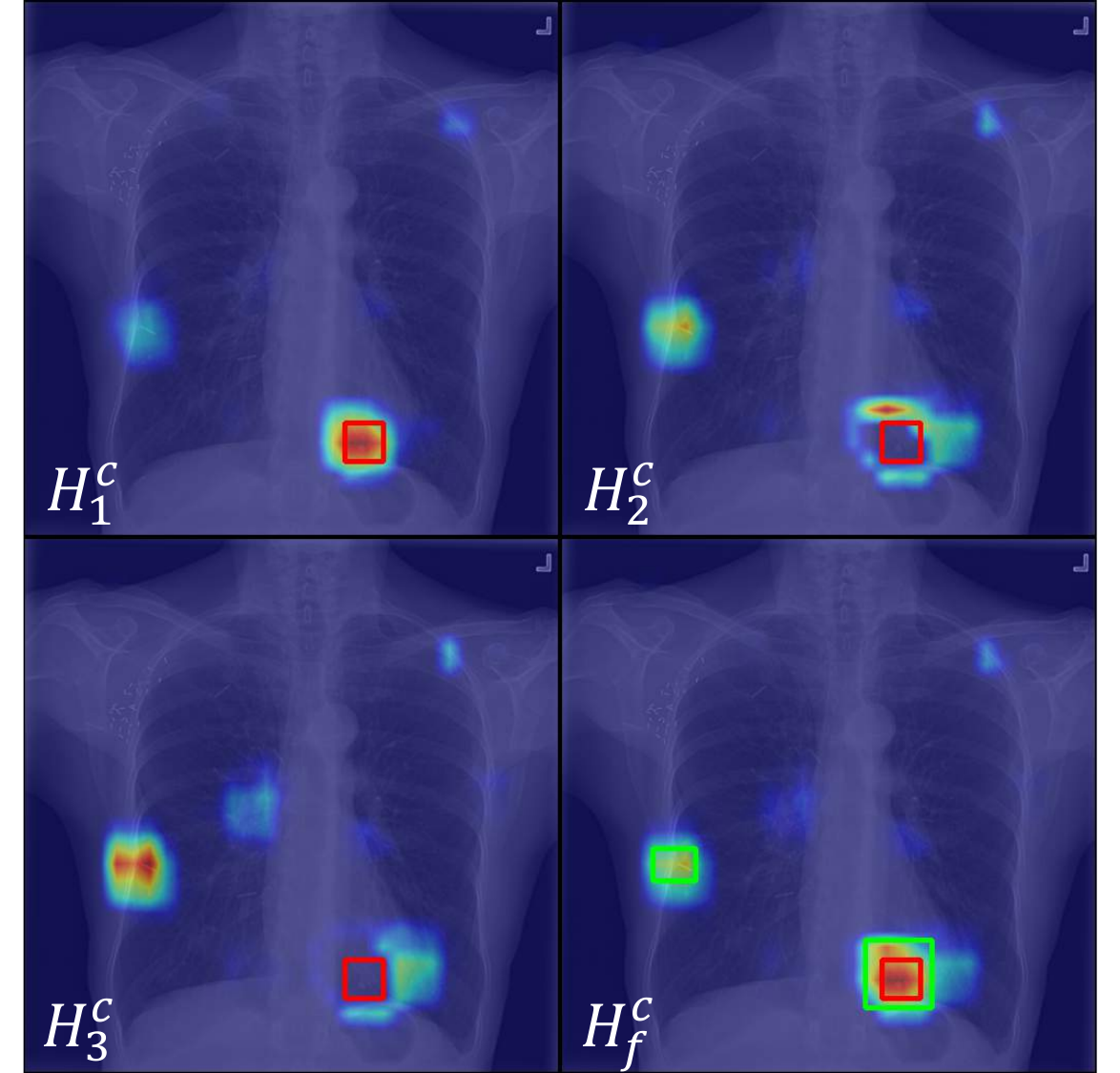}
	\end{subfigure}
	\begin{subfigure}{0.32\textwidth}
		\includegraphics[width=\linewidth, page=18, trim={7cm 0cm 7cm 0cm}, clip]{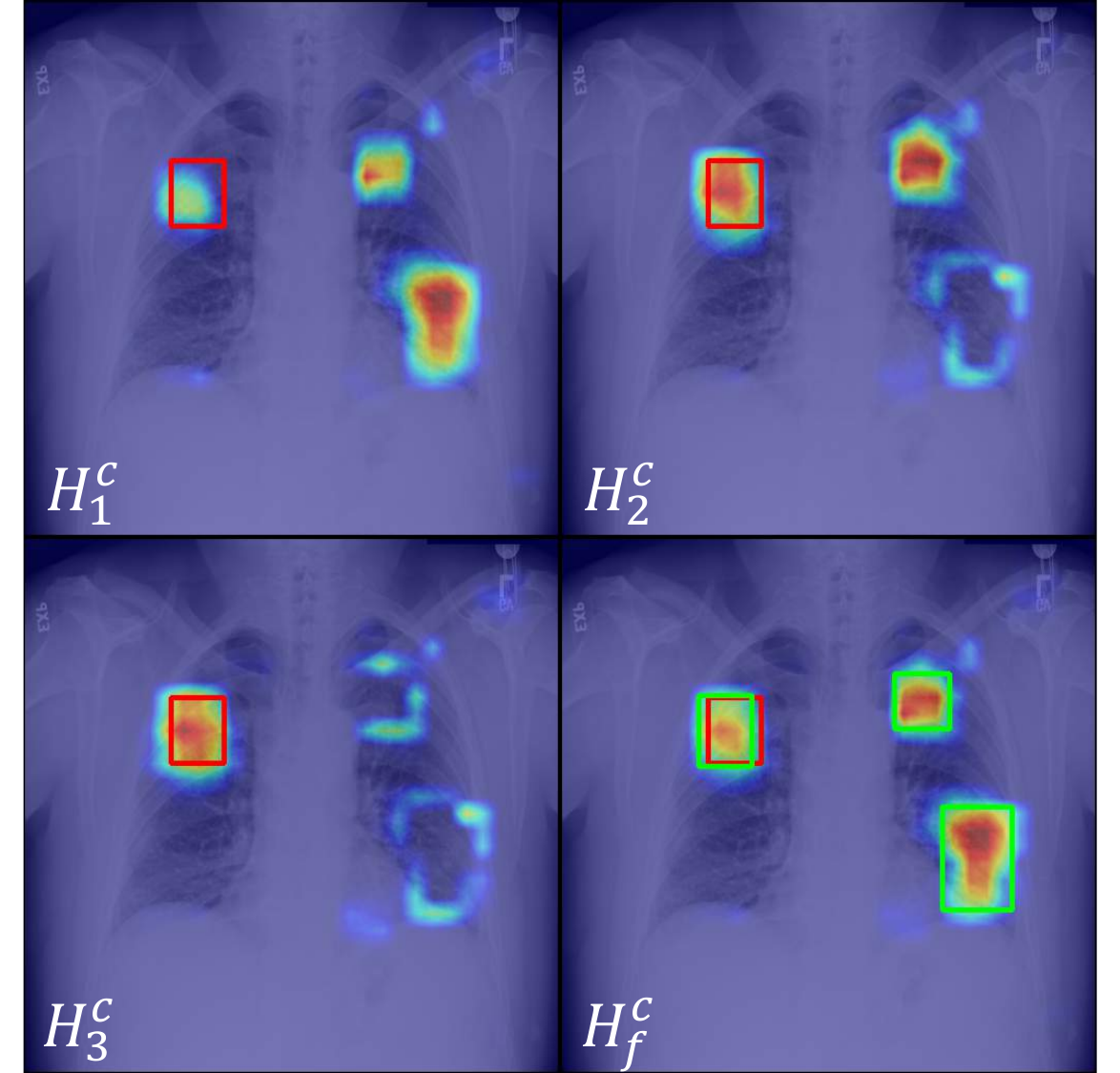}
	\end{subfigure}
	\begin{subfigure}{0.32\textwidth}
		\includegraphics[width=\linewidth, page=19, trim={7cm 0cm 7cm 0cm}, clip]{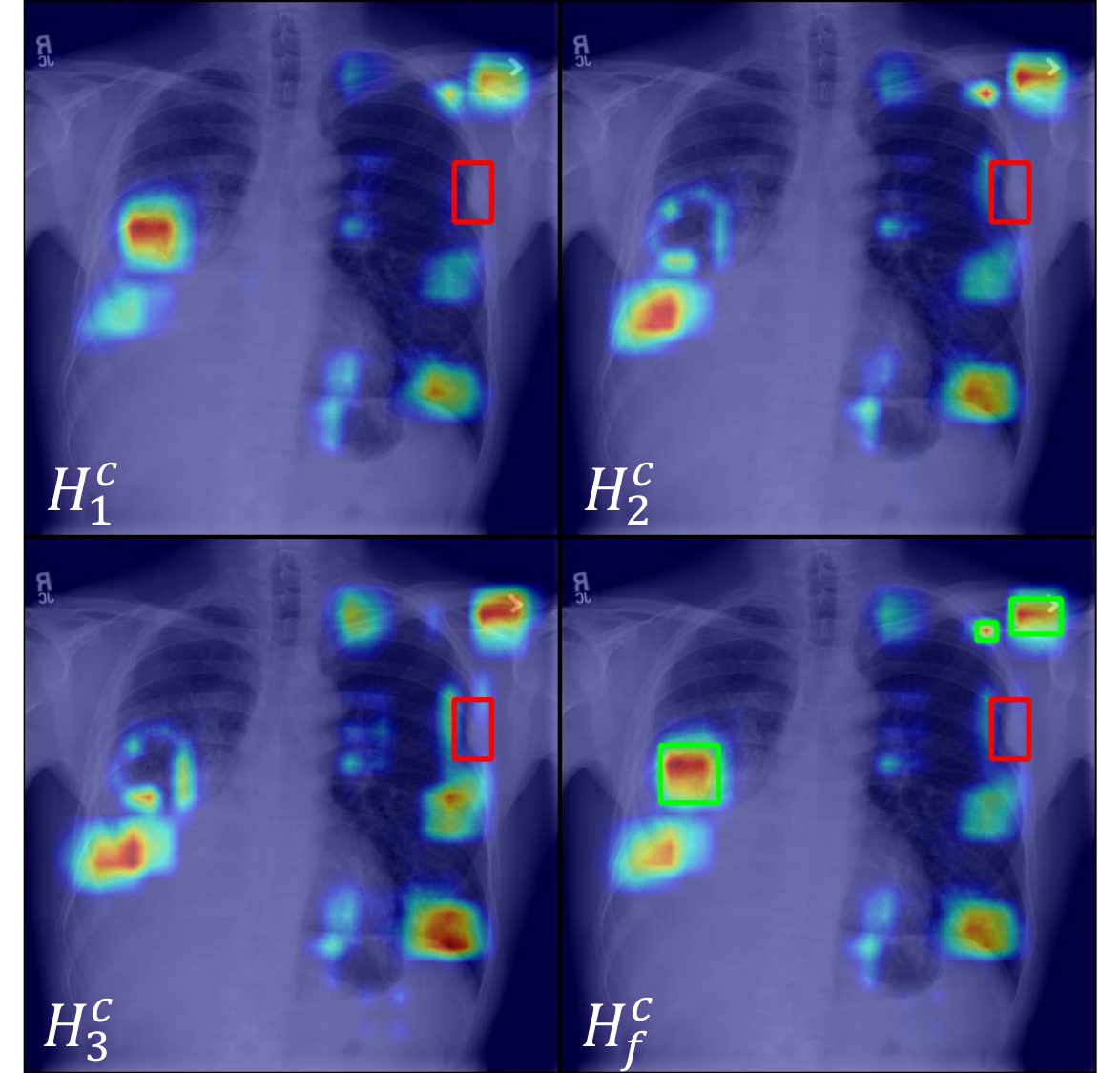}
	\end{subfigure}
	\vspace{0mm}
	\caption{Attetion ming on nodule.}
\end{figure}

\begin{figure}[h!]
	\begin{subfigure}{0.32\textwidth}
		\includegraphics[width=\linewidth, page=20, trim={7cm 0cm 7cm 0cm}, clip]{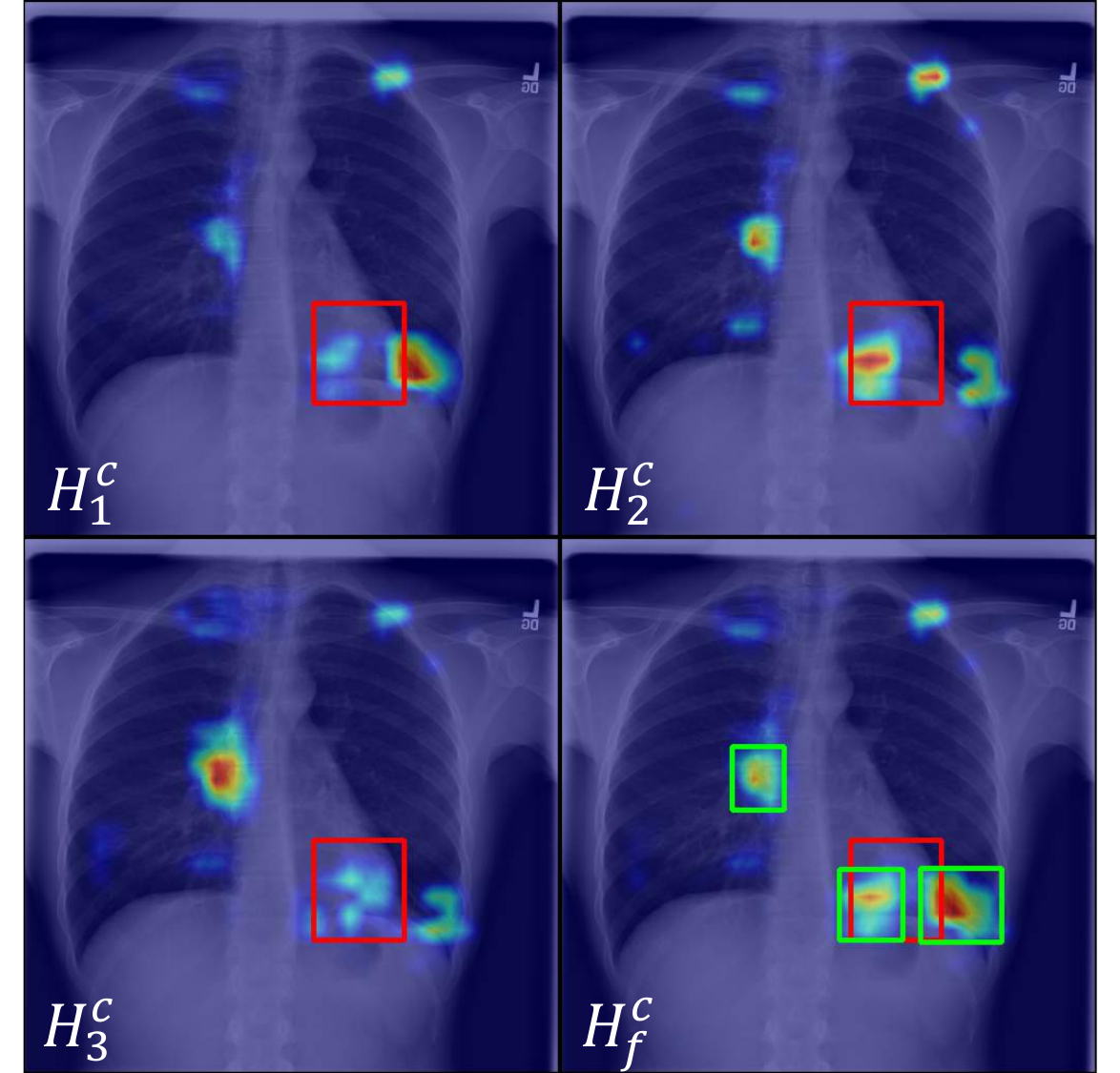}
	\end{subfigure}
	\begin{subfigure}{0.32\textwidth}
		\includegraphics[width=\linewidth, page=21, trim={7cm 0cm 7cm 0cm}, clip]{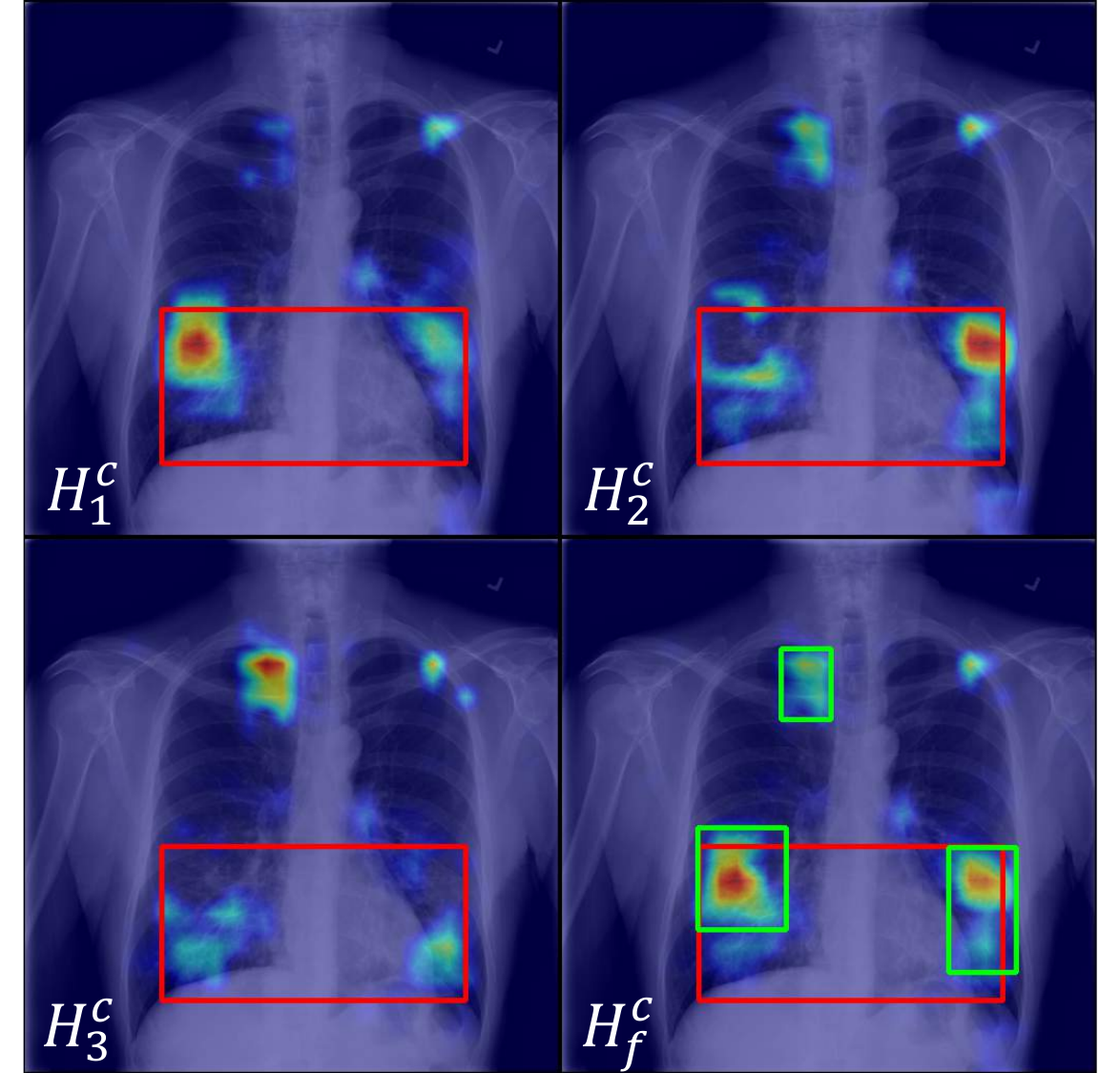}
	\end{subfigure}
	\begin{subfigure}{0.32\textwidth}
		\includegraphics[width=\linewidth, page=22, trim={7cm 0cm 7cm 0cm}, clip]{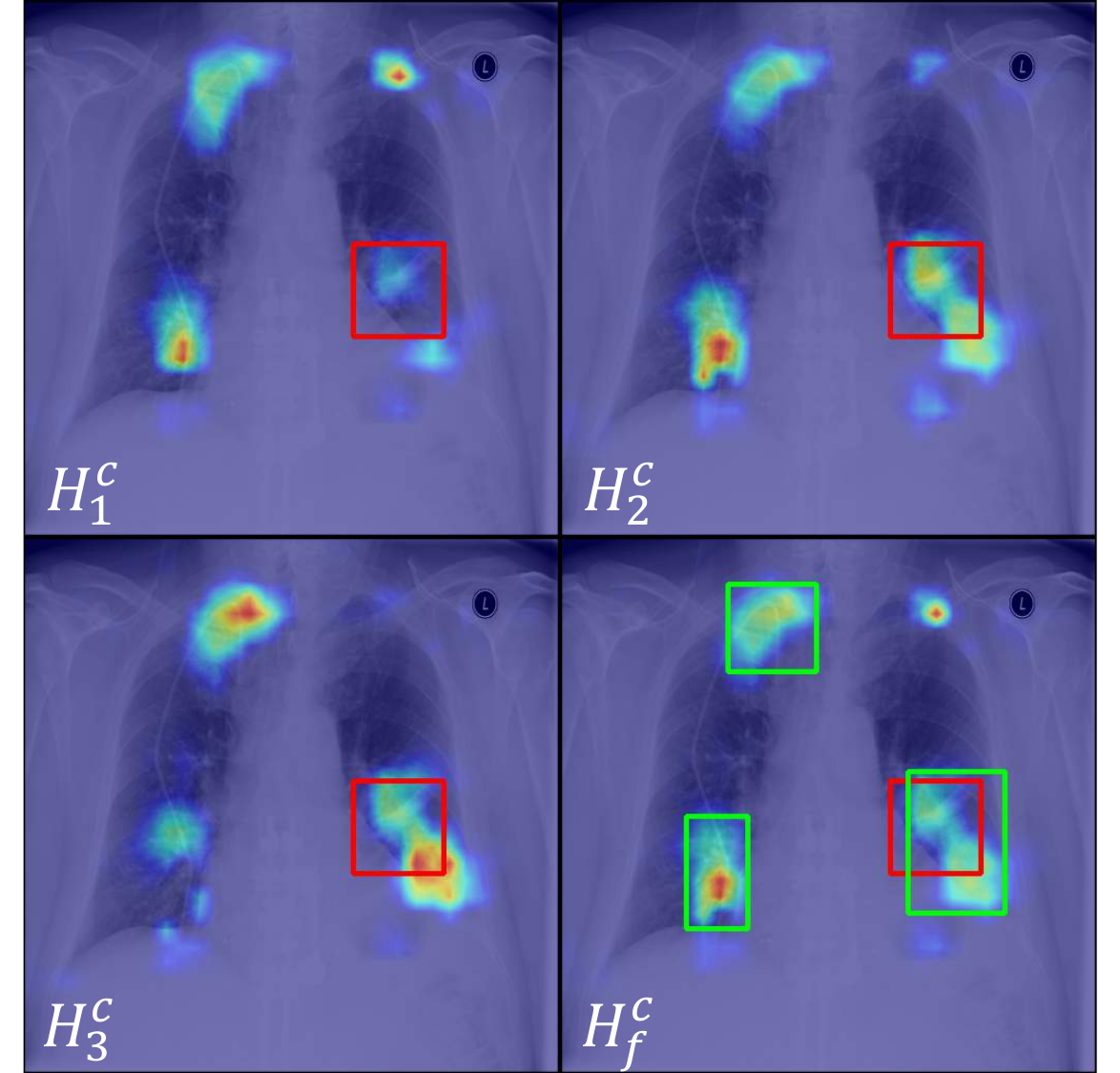}
	\end{subfigure}
	\vspace{0mm}
	\caption{Attetion ming on pneumonia.}
\end{figure}

\begin{figure}[h!]
	\begin{subfigure}{0.32\textwidth}
		\includegraphics[width=\linewidth, page=23, trim={7cm 0cm 7cm 0cm}, clip]{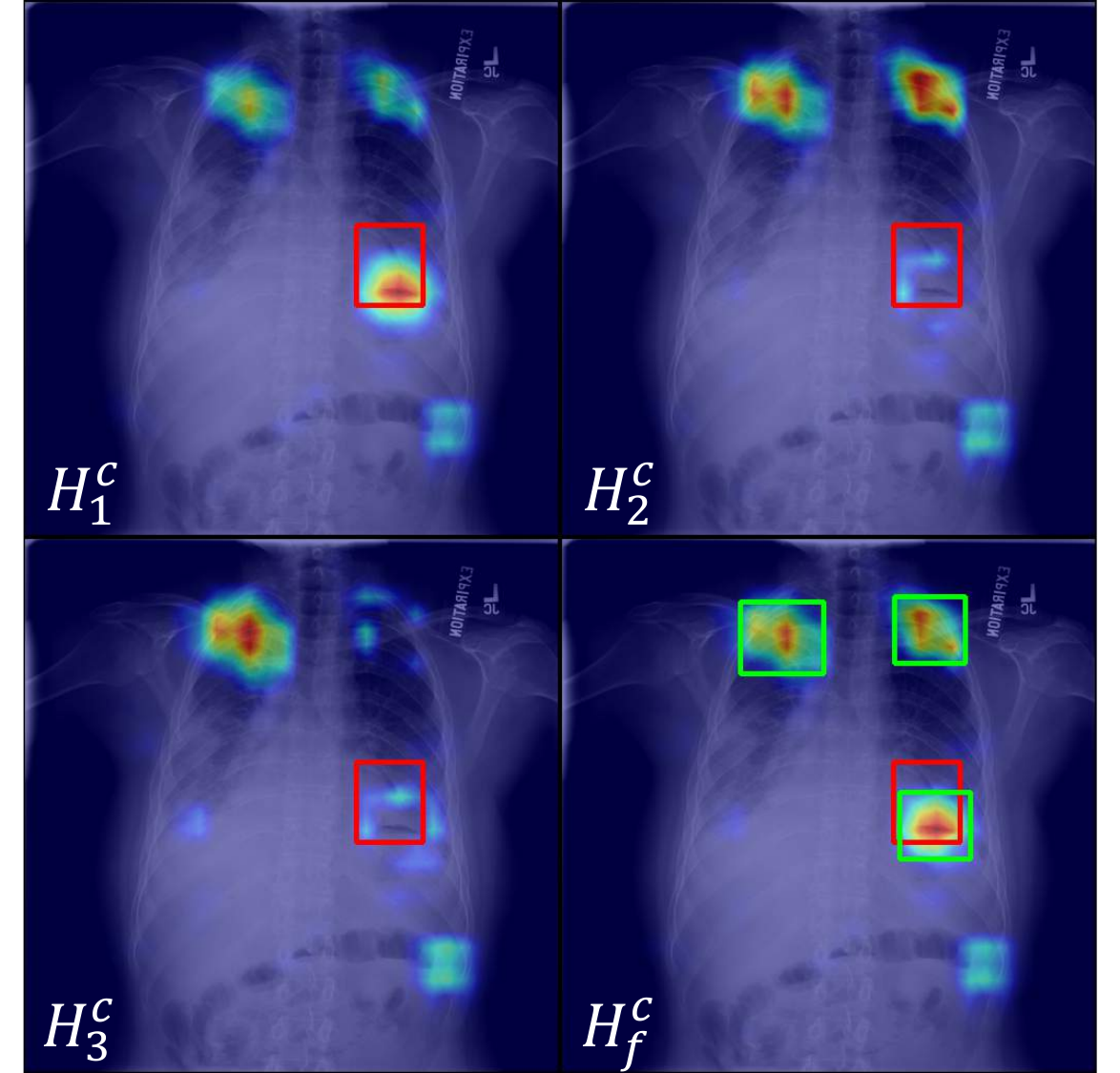}
	\end{subfigure}
	\begin{subfigure}{0.32\textwidth}
		\includegraphics[width=\linewidth, page=24, trim={7cm 0cm 7cm 0cm}, clip]{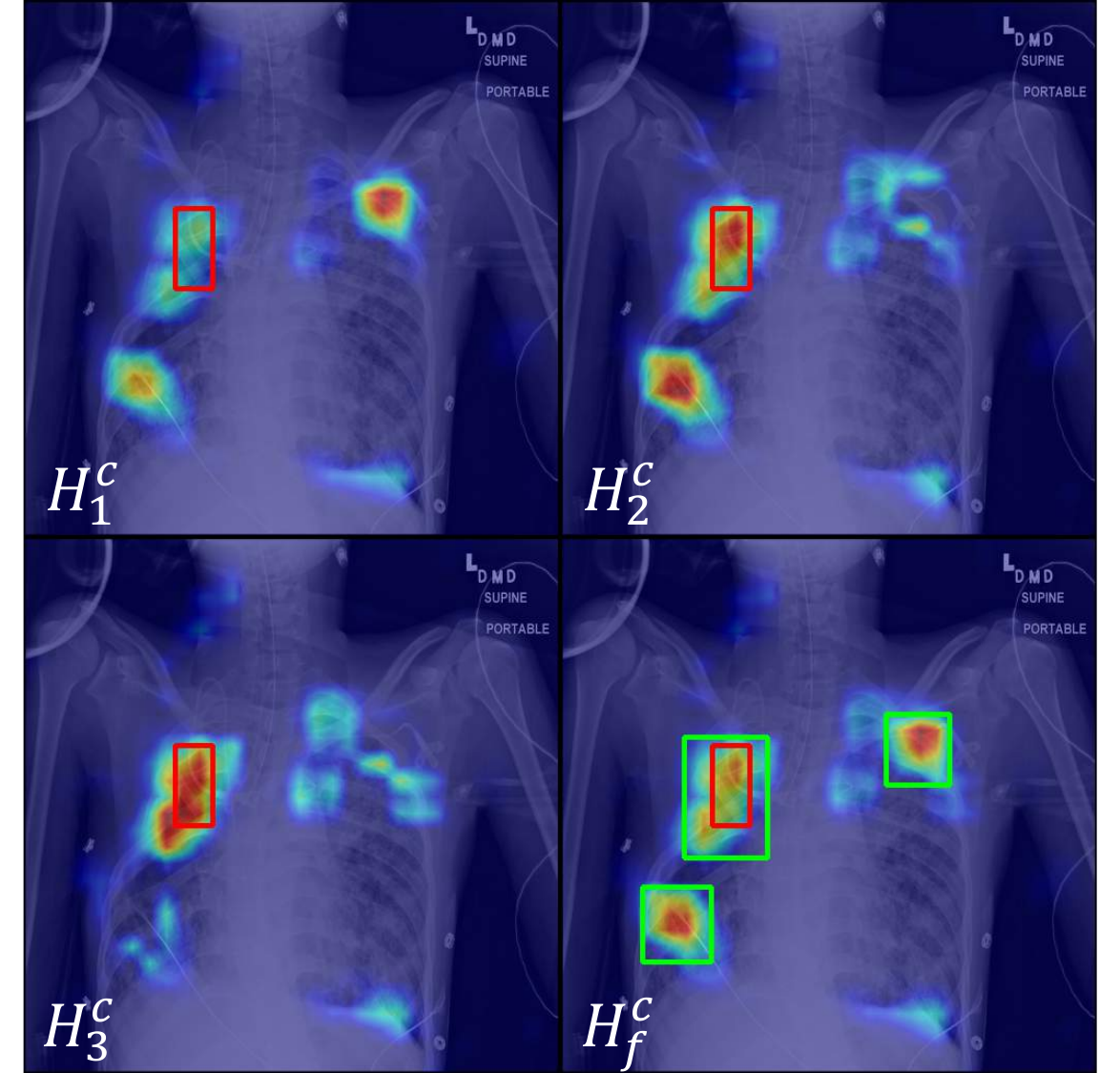}
	\end{subfigure}
	\begin{subfigure}{0.32\textwidth}
		\includegraphics[width=\linewidth, page=25, trim={7cm 0cm 7cm 0cm}, clip]{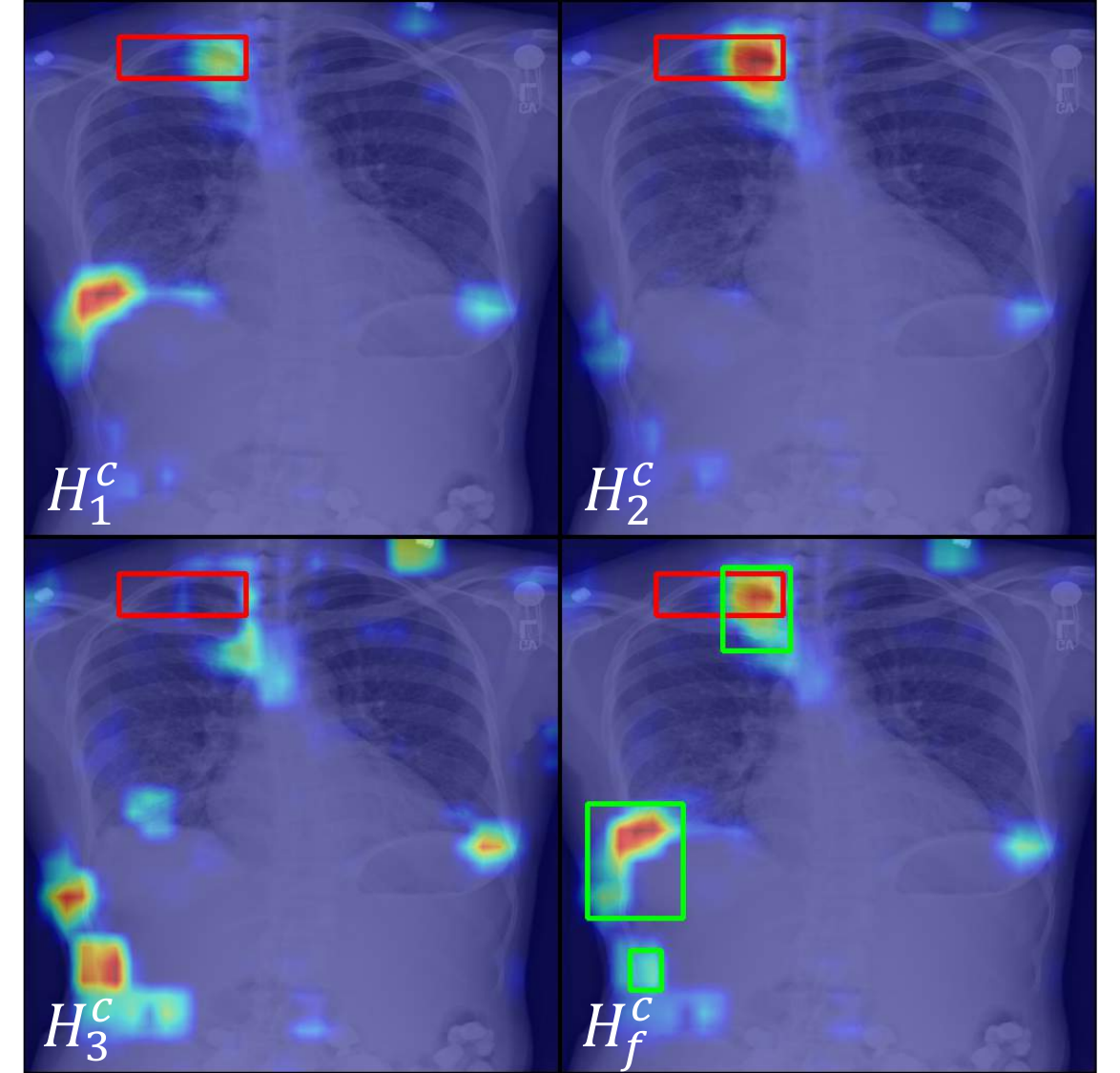}
	\end{subfigure}
	\vspace{0mm}
	\caption{Attetion ming on pneumothorax.}
\end{figure}

\end{document}

%% file: figures/framework.tex

\begin{figure*}[t!]
	\begin{center}
		\includegraphics[width=\linewidth, page=1, trim={2.5cm 4.8cm 5.5cm 2cm}, clip]{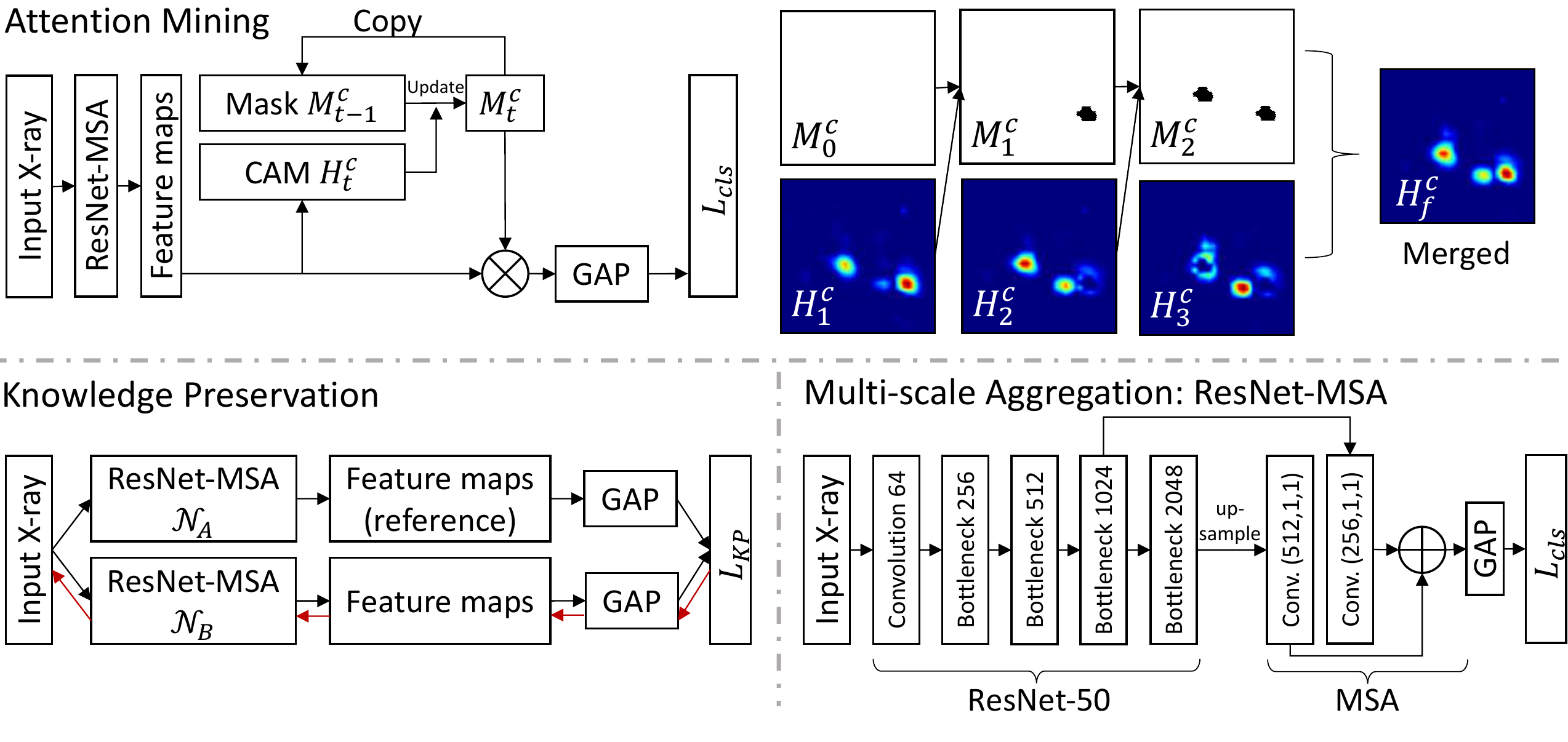}
	\end{center}
	\caption{Architectures of the proposed attention mining (AM), knowledge preservation (KP), and multi-scale aggregation (MSA). Red arrows in the KP module indicate the path of back-propagation. The convolution parameters for MSA are shown as (number of filters, kernel size, stride)
	. See Sec.~\ref{section:method} for details. 
	}
	\label{fig:overview}
\end{figure*}

%% file: data/localization_msa.tex
\begin{table}[t!]
	\centering
	\caption{To compare different MSA setups, each table cell show localization $Acc.$ given T(IoU)=0.3 using all bboxs in $\mathcal{B}$. (See Sec.~\ref{exp:msa} and Sec.~\ref{exp:loc} for details.)}
	\label{tab:msa}
	\scriptsize{
	\begin{tabular}{>{~~}m{2cm} >{\centering}m{1cm} >{\centering}m{1cm} >{\centering}m{1cm} >{\centering}m{1cm} >{\centering}m{1cm} >{\centering}m{1cm} >{\centering}m{1cm} >{\centering\arraybackslash}m{1cm}}
		\toprule
		Method \qquad \qquad & AT & CM & PE & Infiltrat. & Mass & Nodule & PNA & PTx \\
		\midrule
		512-baseline & 0.21 & \textBF{0.81} & \textBF{0.37} & 0.37 & 0.21 & \underline{0.04} & \textBF{0.38} & \textBF{0.35} \\
		512-bilinear & 0.21 & 0.62 & 0.34 & \textBF{0.54} & \textBF{0.35} & \textBF{\underline{0.24}} & 0.37 & 0.29 \\
		512-deconv.   & \textBF{0.28} & 0.55 & 0.35 & 0.50 & 0.32 & 0.20 & 0.35 & 0.27 \\
		1024-baseline& 0.21 & 0.19 & 0.33 & 0.37 & 0.35 & 0.09 & 0.23 & 0.09 \\
		1024-bilinear& 0.11 & 0.10 & 0.30 & 0.30 & 0.15 & 0.22 & 0.05 & 0.08 \\
		1024-deconv.  & 0.07 & 0.10 & 0.23 & 0.40 & 0.13 & 0.01 & 0.05 & 0.03 \\
		\bottomrule
	\end{tabular}
	\vspace{-4mm}
	}
\end{table}

%% file: data/localization_IoU.tex
\begin{table}[t!]
	\centering
	\caption{Comparsion of localization results.  The result ``\#/\#'' is defined as ``our $Acc.$-ref. $Acc.$/our AFP-ref. AFP'', where ``ref.'' is the method for comparison.}

	\label{tab:cmpr}
	\scriptsize{
	\begin{tabular}{l c c c c c c c c}
		\toprule
		T(IoU)& AT & CM & PE & Infiltrat. & Mass & Nodule & PNA & PTx \\ 
		\midrule 
		\multicolumn{9}{c}{Our method compared with baseline (defined in Sec.~\ref{exp:loc})} \\ 
		\midrule
		0.10 & \textBF{0.04/-0.02} & -0.02/-0.02 		 & -0.06/-0.02 		   & \textBF{0.12/-0.02} & \textBF{0.02/-0.01} & \textBF{0.19/-0.01} & \textBF{0.08/-0.02} & -0.11/-0.01 \\
		0.20 & \textBF{0.09/-0.01} & -0.07/-0.01 		 & \textBF{0.04/-0.02} & \textBF{0.20/-0.02} & \textBF{0.14/-0.01} & \textBF{0.27/-0.02} & \textBF{0.09/-0.01} & -0.08/-0.01 \\
		0.30 & \textBF{0.14/-0.02} & -0.09/-0.02 		 & \textBF{0.01/-0.01} & \textBF{0.14/-0.02} & \textBF{0.16/-0.02} & \textBF{0.11/-0.02} & \textBF{0.08/-0.02} & -0.04/-0.01 \\
		0.40 & \textBF{0.15/-0.02} & -0.08/-0.02 		 & \textBF{0.06/-0.02} & \textBF{0.15/-0.01} & \textBF{0.10/-0.02} & \textBF{0.03/-0.02} & \textBF{0.04/-0.02} & -0.01/-0.02 \\
		0.50 & \textBF{0.07/-0.01} & -0.02/-0.02 		 & \textBF{0.04/-0.02} & \textBF{0.08/-0.02} & \textBF{0.02/-0.02} & \textBF{0.03/-0.02} & \textBF{0.01/-0.01} & \textBF{0.00/-0.02} \\
		0.60 & \textBF{0.01/-0.02} & \textBF{0.04/-0.02} & \textBF{0.04/-0.02} & \textBF{0.06/-0.02} & \textBF{0.01/-0.02} & \textBF{0.01/-0.02} & \textBF{0.01/-0.01} & \textBF{0.04/-0.02} \\
		0.70 & \textBF{0.00/-0.02} & -0.02/-0.02 		 & \textBF{0.01/-0.02} & \textBF{0.02/-0.02} & \textBF{0.01/-0.02} & \textBF{0.00/-0.02} & \textBF{0.00/-0.01} & \textBF{0.00/-0.02} \\
		\midrule
		\multicolumn{9}{c}{Our method compared with the reported results in~\cite{wang_2017_chestx}} \\ 
		\midrule
	    0.10 & -0.01/-0.02         & \textBF{0.03/-0.41} & -0.01/-0.02         & -0.19/-0.02         & \textBF{0.16/-0.02} & \textBF{0.32/-0.01} & \textBF{0.02/-0.01} & \textBF{0.05/-0.02} \\
	    0.20 & \textBF{0.03/-0.02} & \textBF{0.22/-0.44} & \textBF{0.07/-0.01} & -0.04/-0.02         & \textBF{0.21/-0.01} & \textBF{0.22/-0.01} & \textBF{0.19/-0.01} & \textBF{0.01/-0.02} \\
	    0.30 & \textBF{0.08/-0.02} & \textBF{0.39/-0.43} & \textBF{0.04/-0.02} & \textBF{0.00/-0.02} & \textBF{0.18/-0.01} & \textBF{0.08/-0.01} & \textBF{0.23/-0.02} & \textBF{0.03/-0.01} \\
	    0.40 & \textBF{0.13/-0.02} & \textBF{0.45/-0.34} & -0.03/-0.02         & \textBF{0.07/-0.02} & \textBF{0.11/-0.01} & \textBF{0.01/-0.01} & \textBF{0.15/-0.02} & \textBF{0.04/-0.01} \\
	    0.50 & \textBF{0.06/-0.02} & \textBF{0.42/-0.24} & -0.01/-0.02         & \textBF{0.06/-0.02} & \textBF{0.06/-0.02} & \textBF{0.01/-0.01} & \textBF{0.14/-0.01} & \textBF{0.05/-0.02} \\
	    0.60 & \textBF{0.01/-0.02} & \textBF{0.36/-0.09} & \textBF{0.00/-0.02} & \textBF{0.03/-0.02} & \textBF{0.05/-0.02} & \textBF{0.00/-0.01} & \textBF{0.03/-0.02} & \textBF{0.04/-0.02} \\
	    0.70 & \textBF{0.00/-0.02} & \textBF{0.14/-0.02} & -0.01/-0.02         & \textBF{0.02/-0.02} & \textBF{0.01/-0.02} & \textBF{0.00/-0.01} & \textBF{0.01/-0.02} & \textBF{0.00/-0.02} \\
		\bottomrule
	\end{tabular}
	}
\end{table}

%% file: figures/show_AM.tex

\begin{figure}[t!]
	\begin{subfigure}{0.32\textwidth}
		\includegraphics[width=\linewidth, page=2, trim={7cm 0cm 7cm 0cm}, clip]{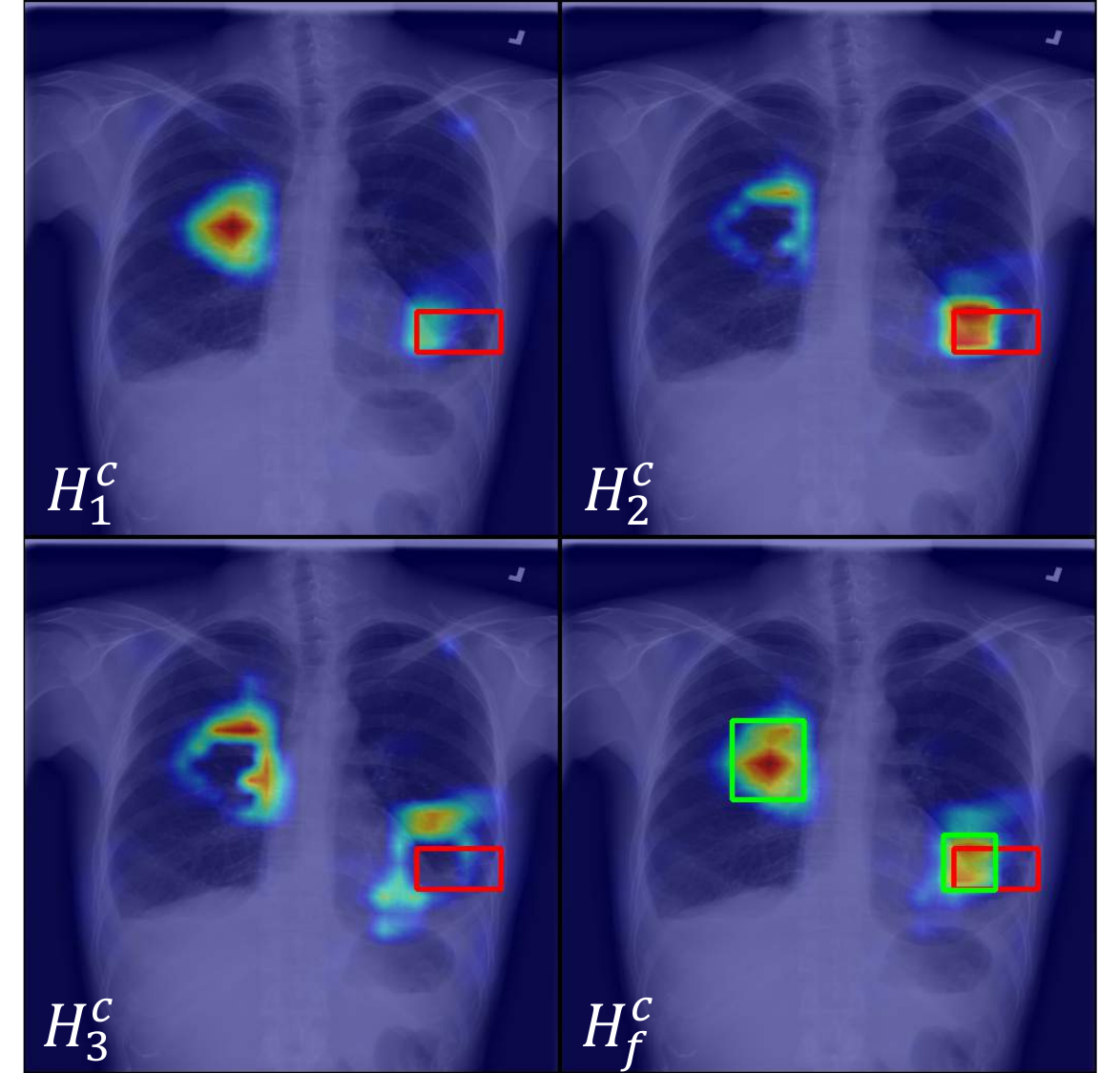}
		\caption{object mined at $t=2$}
	\end{subfigure}
	\begin{subfigure}{0.32\textwidth}
		\includegraphics[width=\linewidth, page=3, trim={7cm 0cm 7cm 0cm}, clip]{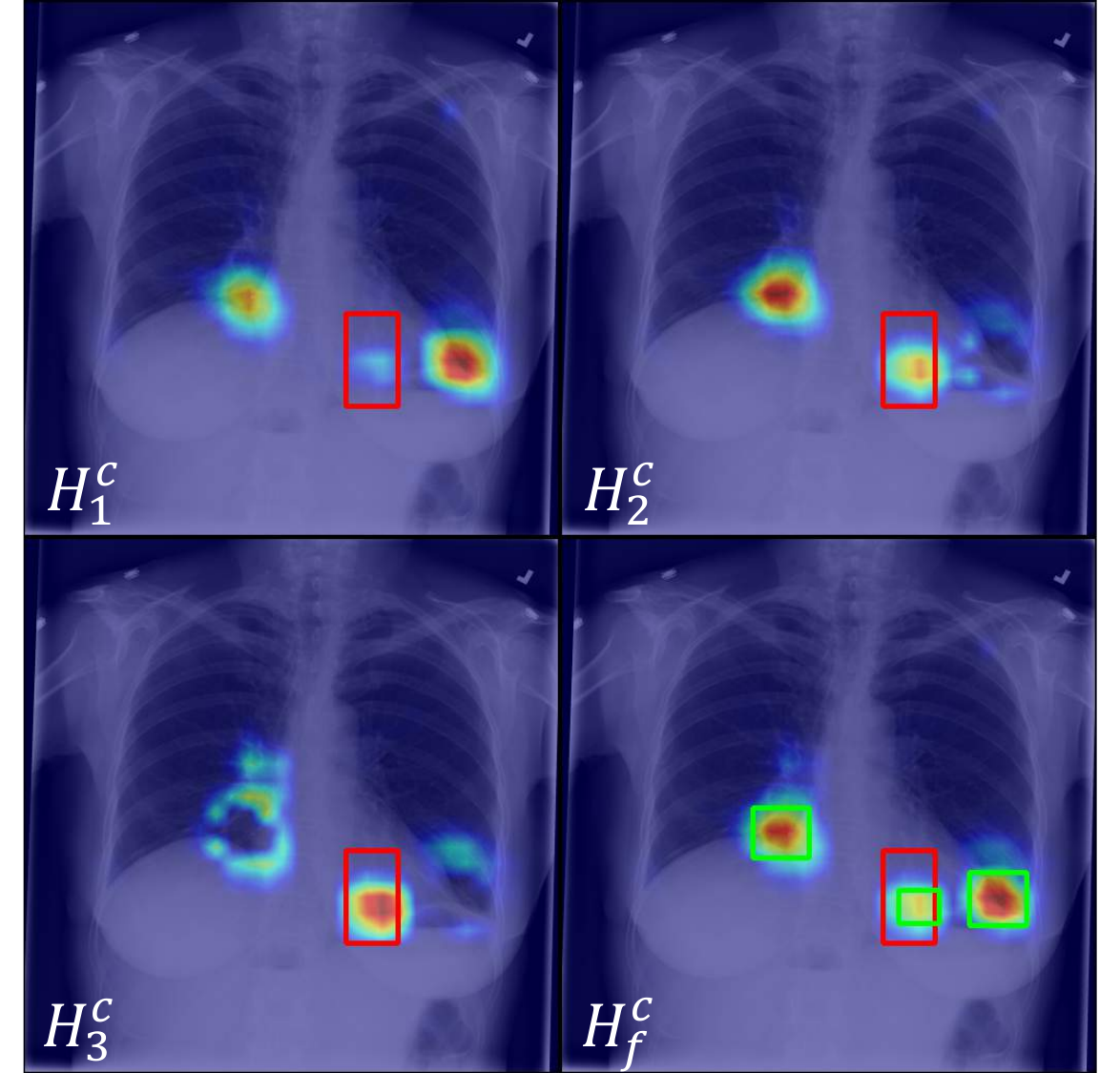}
		\caption{object mined at $t=3$}
	\end{subfigure}
	\begin{subfigure}{0.32\textwidth}
		\includegraphics[width=\linewidth, page=4, trim={7cm 0cm 7cm 0cm}, clip]{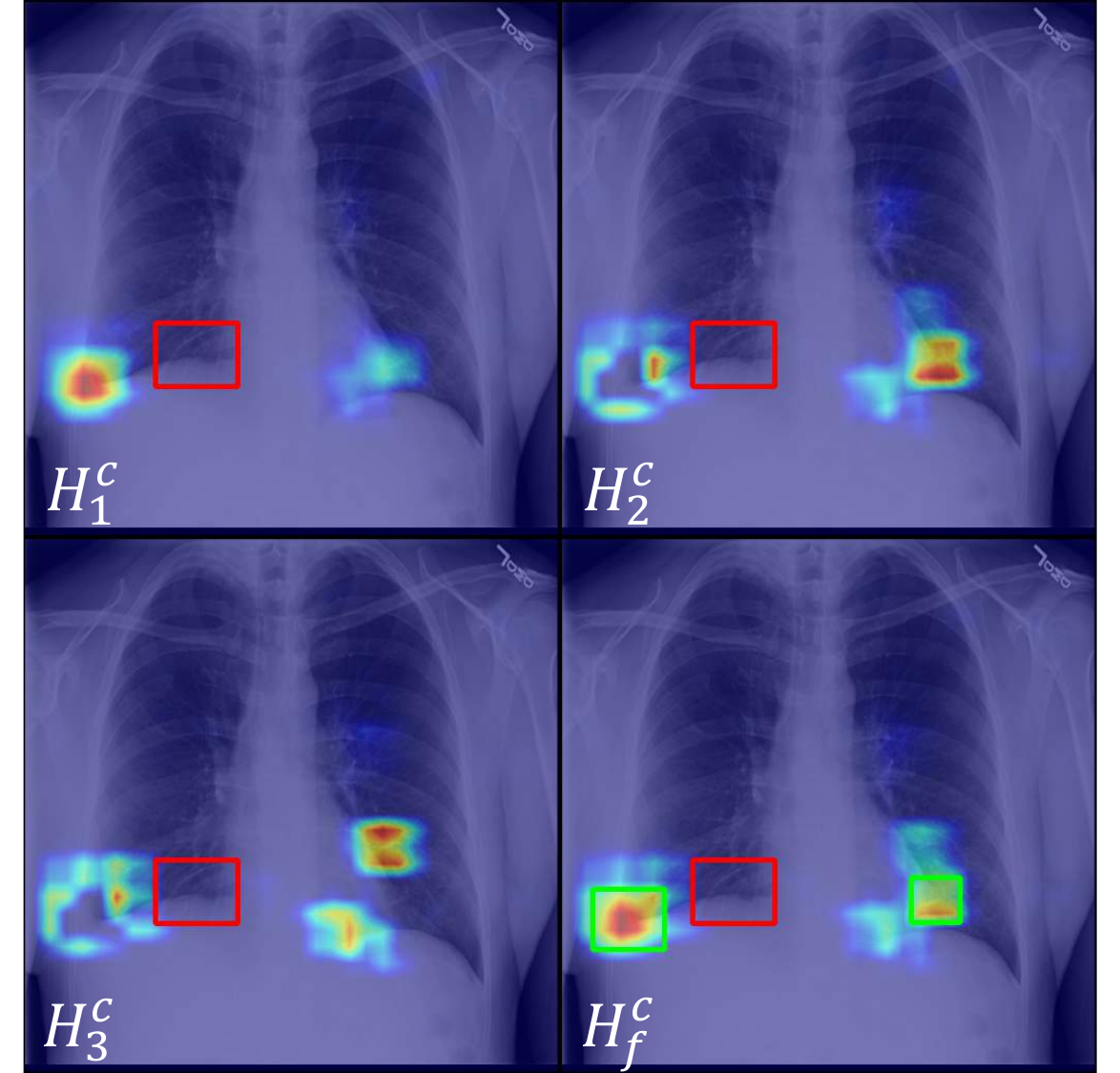}
		\caption{the failure case}
	\end{subfigure}
	\caption{Visualization of heatmaps generated during attention mining. The ground truth and the automatic bboxs  are colored in red and green, respcetively. }
	\label{fig:show-am}
\end{figure}

%% file: data/localization_ablation.tex
\begin{table}[t!]
	\centering
	\caption{Ablation study of attention mining (AM) and knowledge preservation (KP). Each table cell shows the $Acc.$ by using all of the bboxs in $\mathcal{B}$ (in Sec.~\ref{exp:loc}). }
	\label{tab:ablation}
	\scriptsize{
	\begin{tabular}{>{\centering}m{0.5cm}>{\centering}m{1cm}>{\centering}m{1.5cm} >{\centering}m{1cm} >{\centering}m{1cm} >{\centering}m{1cm} >{\centering}m{1cm} >{\centering}m{1cm} >{\centering}m{1cm} >{\centering}m{1cm} >{\centering\arraybackslash}m{1cm}}
		\toprule
		& T(IoU) & Method & AT & CM & PE & Infiltrat. & Mass & Nodule & PNA & PTx \\ 
		\midrule 
		
		\multirow{6}{*}{AM}
		& \multirow{3}{*}{0.1}
		& t=1 & 0.57 		  & 0.96 		  & \textBF{0.84} & 0.78 		 & 0.58 		 & \textBF{0.65} & 0.66 		 & 0.68 \\
		
		& & t=2 & 0.65 		  & \textBF{0.97} & 0.82 		  & 0.77 		 & 0.60 		 & 0.61 		 & 0.72 		 &\textBF{0.72} \\
		& & t=3 & \textBF{0.68} & \textBF{0.97} & 0.83 		  &\textBF{0.79} & \textBF{0.62} & 0.57 		 & \textBF{0.73} &\textBF{0.72} \\ 
		\cmidrule{2-11}
		& \multirow{3}{*}{0.3}
		& t=1 & \textBF{0.34} & 0.73 		  & 0.47 		  & 0.40 		 &\textBF{0.36} &\textBF{0.33} & 0.33 		   & 0.30 \\
		& & t=2 & 0.32 		  & 0.82 		  & 0.47 		  & 0.40 		 & 0.34 		& 0.25 		   & 0.42 		   & 0.36 \\
		& & t=3 & 0.33 		  & \textBF{0.85} & \textBF{0.48} &\textBF{0.42} & 0.34 		& 0.20 		   & \textBF{0.44} &\textBF{0.38} \\
		
		\midrule
		
		\multirow{6}{*}{KP}
		& \multirow{3}{*}{0.1}
		& w/o KP 	 & 0.67 		  & \textBF{1.00} & 0.48 		  & 0.58 		  & 0.54 		  & 0.47 		  & 0.70 		  & 0.71 \\ 
		& & KP-Vanilla & 0.65 		  & 0.97 		  & 0.78 		  & \textBF{0.87} & 0.61 		  & 0.48 		  & \textBF{0.73} & \textBF{0.72} \\
		& & KP 	     & \textBF{0.68}  & 0.97 		  & \textBF{0.83} & 0.79 		  & \textBF{0.62} & \textBF{0.57} & \textBF{0.73} & \textBF{0.72} \\ 
		\cmidrule{2-11}
		& \multirow{3}{*}{0.3}
		& w/o KP     & 0.22 		 & \textBF{0.99} & 0.11 		 & 0.20 		 & 0.20 		 & 0.06 		 & 0.29 		 & 0.34 \\
		& & KP-Vanilla & 0.26 		 & 0.73 		 & 0.41 		 & \textBF{0.46} & 0.32 		 & 0.10 		 & 0.42 		 & 0.36 \\
		& & KP     	 & \textBF{0.33} & 0.85 		 & \textBF{0.48} & 0.42 		 & \textBF{0.34} & \textBF{0.20} & \textBF{0.44} & \textBF{0.38} \\ 
		\bottomrule

	\end{tabular}
	}
	\vspace{-6mm}
\end{table}